\documentclass[runningheads]{llncs}

\usepackage{eccv}

\usepackage{shortcuts}
\usepackage{multirow}
\newcommand{\Skip}[1]{}

\usepackage{amsthm}
\usepackage{wrapfig}
\usepackage{caption}
\usepackage{subcaption}
\usepackage{graphicx}
\usepackage{xcolor}
\usepackage{colortbl}
\usepackage{makecell}
\definecolor{ourcolor}{RGB}{230,240,255}  

\usepackage{eccvabbrv}

\usepackage{graphicx}
\usepackage{booktabs}

\usepackage[accsupp]{axessibility}  

\usepackage{hyperref}

\usepackage{orcidlink}

\begin{document}

\title{RoME: Robust Mixture of Low-Rank Experts against Multiple Adversarial Perturbations} 

\titlerunning{RoME: Robust Mixture of Low-Rank Experts}

\author{Woo Jae Kim\inst{1} \and
Kyle Min\inst{2} \and
Suhyeon Ha\inst{1} \and
Joonsung Jeon\inst{1} \and
Sung-eui Yoon\inst{1}}

\authorrunning{W.~Kim \etal}

\institute{KAIST, Daejeon, Korea \and
Oracle, Seattle WA, USA \\
\email{\{wkim97, suhyeon.ha, mikeraph\}@kaist.ac.kr, sungeui@kaist.edu}\\
\email{kyle.min@oracle.com}
}

\maketitle

\begin{abstract}
Multi-perturbation adversarial training (MAT) aims to achieve robustness against multiple $\ell_p$ perturbations but suffers from robustness trade-offs between different threats. To address this, we employ a mixture of experts (MoE) to route different threats through distinct model pathways. However, na\"ive application of MoE encounters two critical challenges: experts tend to overlook threat-specific features and redundantly capture features shared across threats, and gating networks suffer from \emph{threat-agnostic routing} where they learn nearly identical routing patterns across threats, thus preventing the construction of threat-specific model pathways. To this end, we propose Robust Mixture of Low-Rank Experts (RoME), where each expert is a low-rank additive update to the shared backbone, allowing it to capture threat-common features while experts focus on threat-specific information. To address threat-agnostic routing, RoME introduces (i) dual-scale gating that exploits threat-discriminative signals from local and global level features, and (ii) threat-guided gating diversification that enforces diverse expert utilization across threats. Extensive experiments demonstrate that RoME outperforms existing state-of-the-art MAT in union robustness and natural accuracy and improves robustness against unseen threats. Codes are available at \url{https://github.com/wkim97/RoME}.
\keywords{Multi-Perturbation Adversarial Training \and Mixture of Experts \and Adversarial Robustness}
\end{abstract}    
\begin{figure}[t] 
    \centering
    \begin{minipage}[t]{0.35\textwidth} 
        \centering
        \vspace{0pt} 
        
        \begin{subfigure}{\textwidth}
            \centering
            \includegraphics[width=0.48\textwidth]{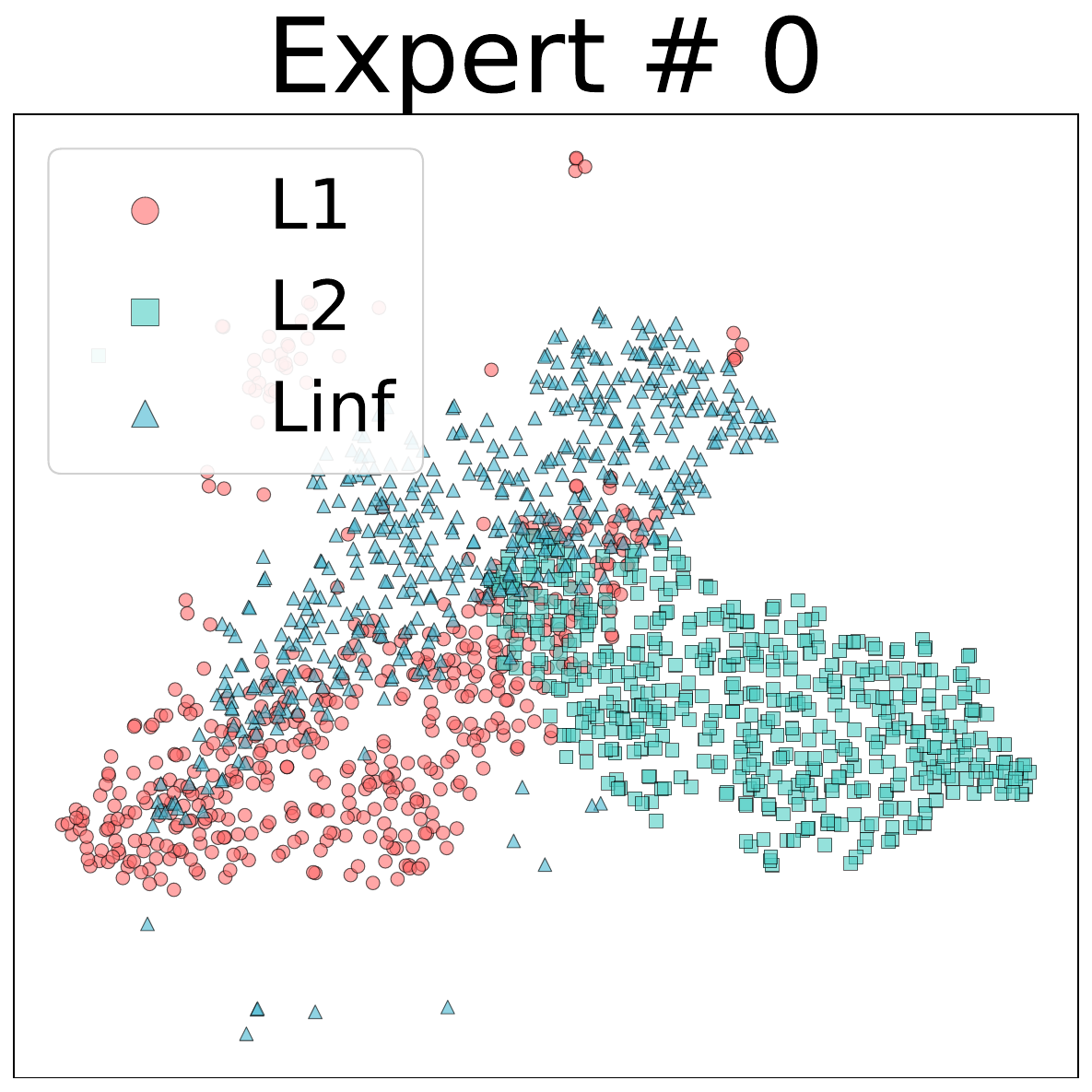}
            \hfill
            \includegraphics[width=0.48\textwidth]{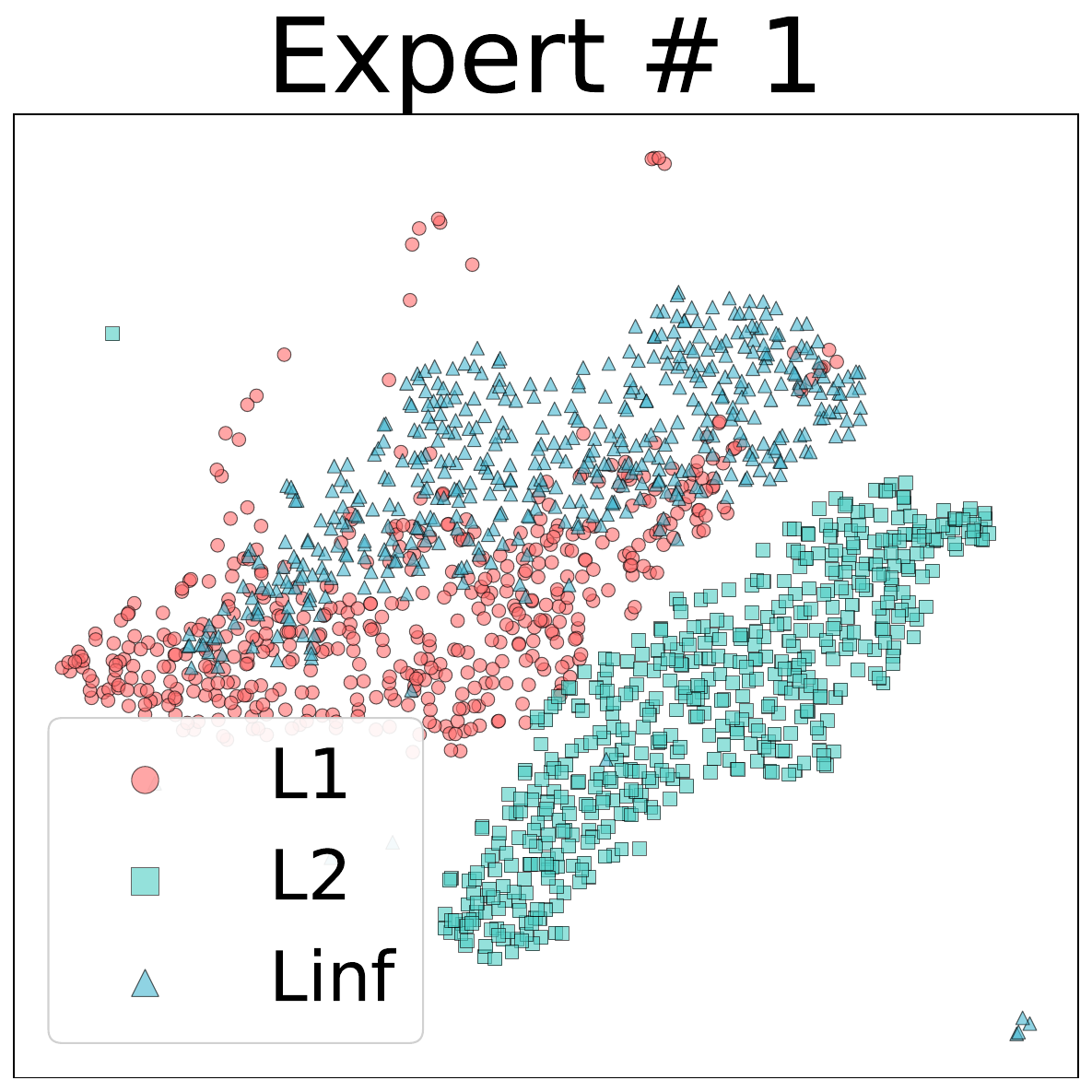}
            \caption{Conventional MoE}
            \label{fig:tsne_baseline}
        \end{subfigure}
        
        \vspace{0pt}
        
        \begin{subfigure}{\textwidth}
            \centering
            \includegraphics[width=0.48\textwidth]{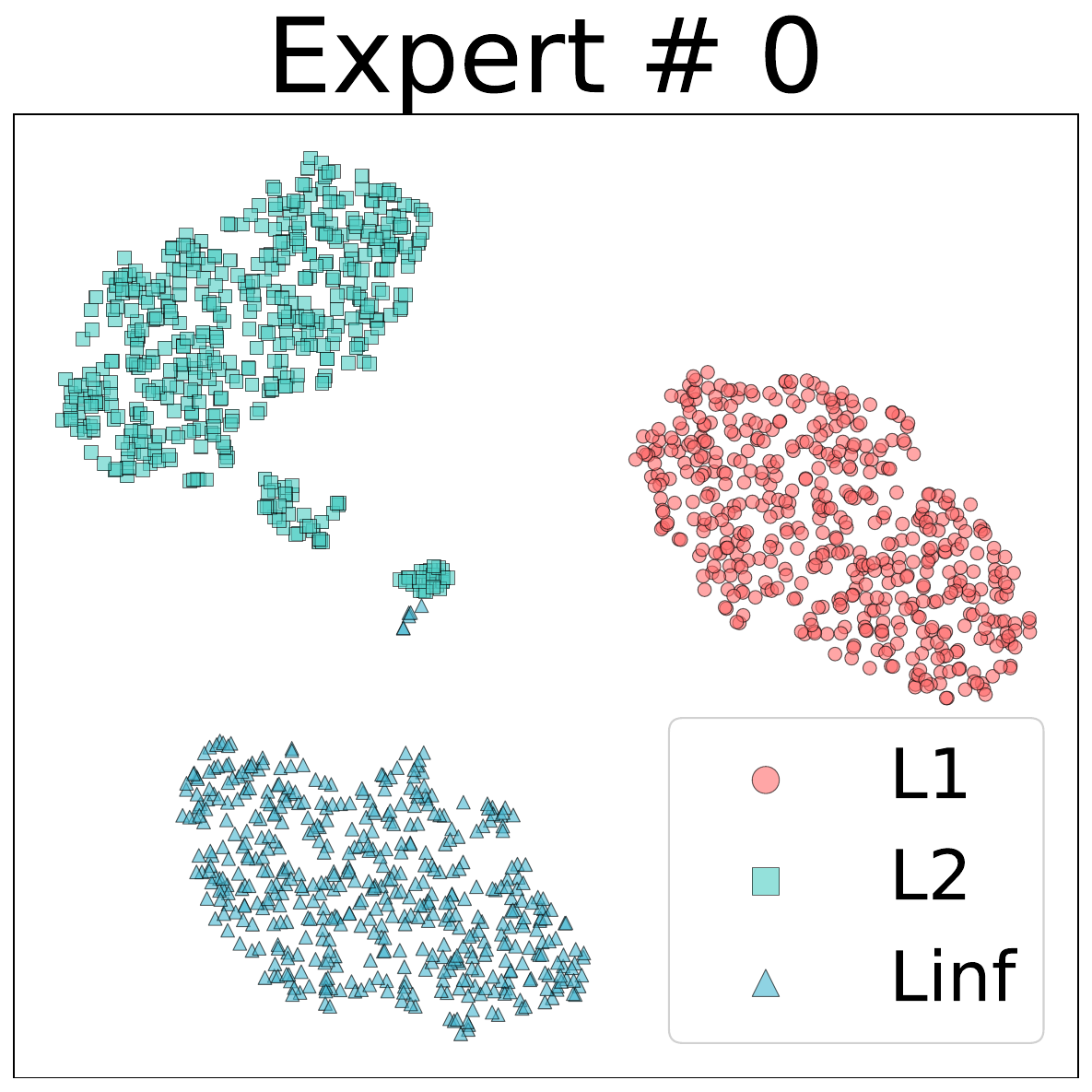}
            \hfill
            \includegraphics[width=0.48\textwidth]{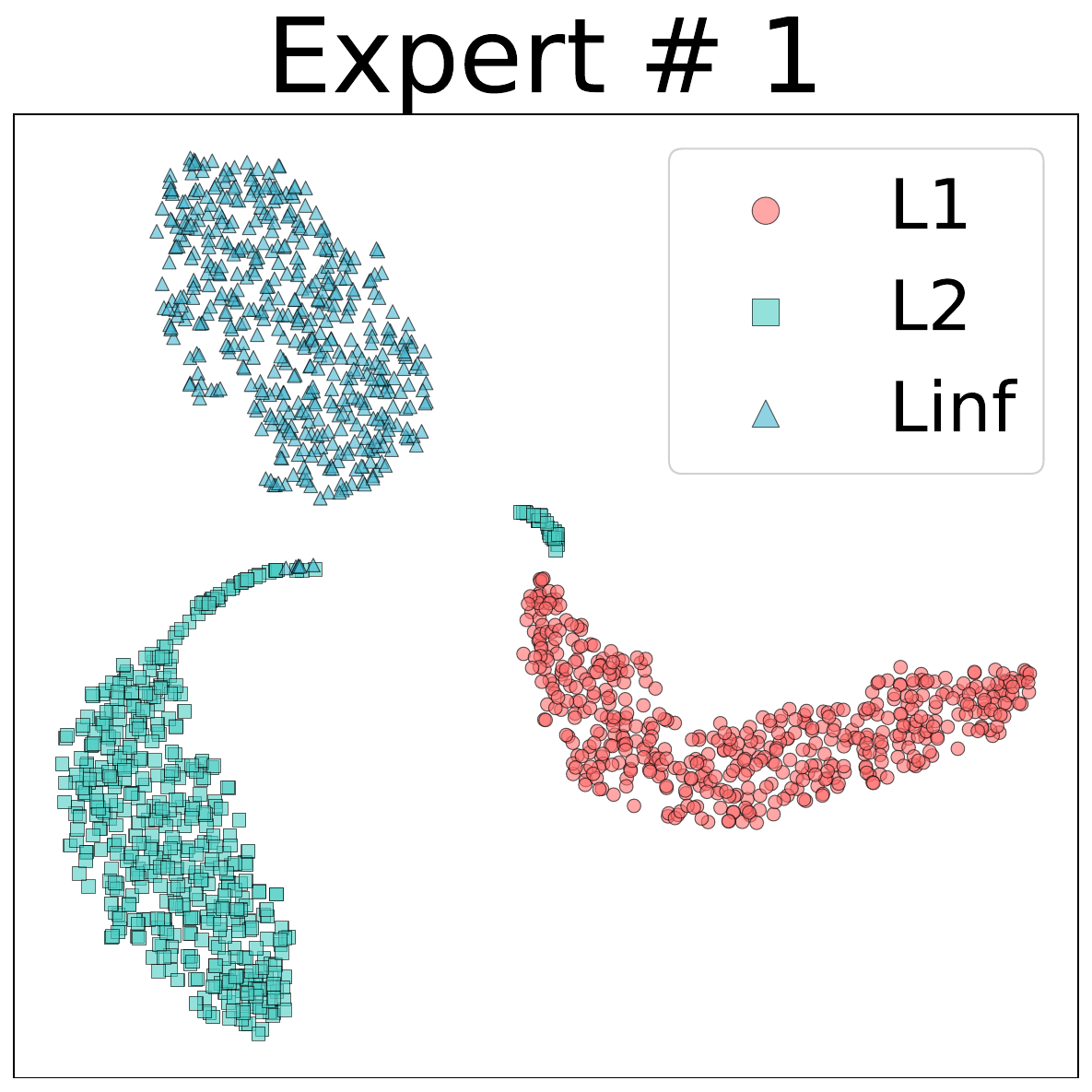}
            \caption{Ours (Low-rank experts)}
            \label{fig:tsne_ours}
        \end{subfigure}

        \caption{
            t-SNE visualization of expert outputs.
            (a) In conventional MoE, disjoint FFN experts redundantly capture similar \textbf{threat-common features}, while (b) our low-rank experts effectively capture \textbf{threat-specific features}.
        }
        \label{fig:teaser_tsne} 
    \end{minipage}
    \hfill
    \begin{minipage}[t]{0.6\textwidth} 
        \centering
        \vspace{0pt} 
        \includegraphics[width=\textwidth]{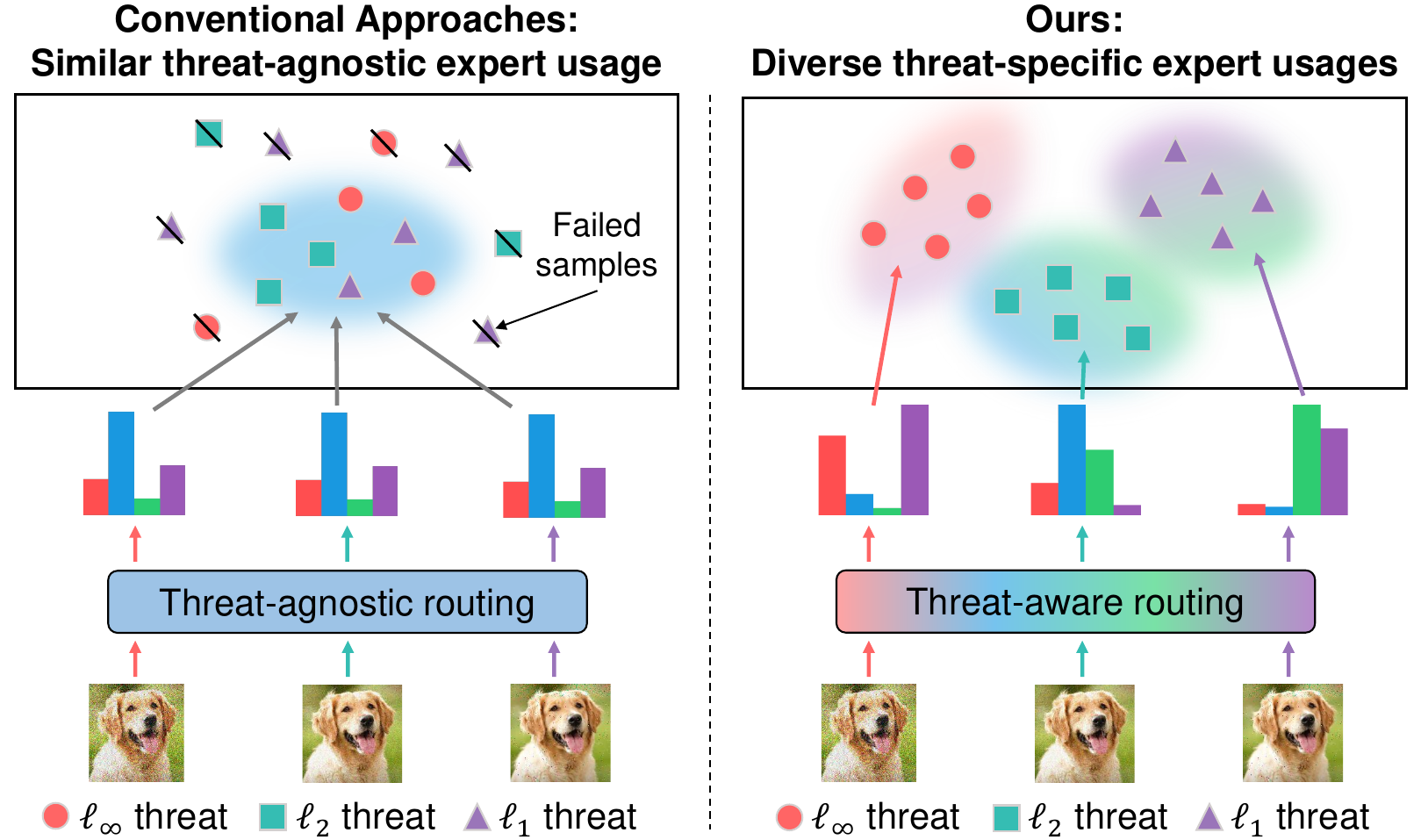}
        \vspace{-5pt}
        \caption{
            Conceptual visualization of conventional mixture of experts and our approach.
            Conventional methods (left) face the \textbf{threat-agnostic routing} issue and route different threats through similar expert combinations, resulting in a single shared model pathway for distinct threats.
            Our approach (right) effectively routes each threat through distinct expert combinations, constructing multiple threat-specific model pathways.
        }
        \label{fig:teaser_concept} 
    \end{minipage}
    \vspace{-10pt}
\end{figure}

\section{Introduction}
\label{sec:intro}
Adversarial training~\cite{fgsm, pgd} has become one of the most effective defenses against adversarial threats.
However, traditional adversarial training typically focuses on a single type of threat (\eg, $\ell_\infty$ perturbations), remaining vulnerable to unseen threats during inference~\cite{max, msd}.
Multi-perturbation adversarial training (MAT)~\cite{max, msd, sat, e-at, ramp} addresses this by training on diverse $\ell_p$ threats simultaneously, improving robustness across multiple threats.

Despite its effectiveness, MAT faces robustness trade-offs when learning on diverse threats, suffering from suboptimal robustness on individual threat types~\cite{max, e-at, ramp}.
This degradation arises because different adversarial threats induce distinct distributional shifts, creating conflicting optimization objectives during training~\cite{e-at, ramp}.
Existing approaches~\cite{max, e-at, ramp} force these distinct threats through a single, fixed architectural pathway, where feature representations learned to be robust against one threat may fail to transfer robustness to others.

To address this, we employ mixture of experts (MoE)~\cite{deepseekmoe, qwen3, shazeer2017outrageously, switch-transformers} to route different threats through distinct model pathways, each learning representations robust to a different type of threat.
However, na\"ive application of MoE~\cite{moe-survey, shazeer2017outrageously, switch-transformers} encounters two critical challenges.
First, conventional experts, typically implemented as multiple independent FFN layers, struggle to capture features unique to each threat.
Adversarial examples crafted from an image naturally share the same underlying image content~\cite{not-bugs, humans-decipher}, and certain threats even share overlapping features~\cite{e-at} (\eg, $\ell_1$ and $\ell_\infty$ robustness partially transfer to $\ell_2$).
Without a shared backbone to separate threat-common from threat-specific features~\cite{mtl-lora}, conventional experts redundantly capture the former as shown in Fig.~\ref{fig:tsne_baseline}.

The shared features among threats further raise a second challenge of \emph{threat-agnostic routing}, where the gating network routes different threats to similar expert combinations as shown in Fig.~\ref{fig:teaser_concept}, failing to construct distinct pathways for each threat.
This occurs because the overlapping features across threat types provide insufficient discriminative signal, preventing the gating network from distinguishing one threat from another.
This forces a single model pathway to learn robust representations against multiple threats, facing the same multi-threat trade-off in existing MAT methods.

To address these challenges, our key insight is two-fold.
First, inspired by a recent approach~\cite{mtl-lora} in multi-task learning that handles different tasks using low-rank experts on top of a shared backbone, we shift from conventional disjoint experts and implement each expert as a \textbf{low-rank additive update} to the shared backbone weights.
This allows the backbone to capture threat-common features, while each expert focuses on learning threat-specific information.

To address threat-agnostic routing, we make an observation that different threats exhibit discriminative cues at different levels of granularity (Fig.~\ref{fig:tsne_dual} in Sec.~\ref{ssec:bilevel_gating}); $\ell_1$ threats are more distinguishable at local patch-level features due to their sparse perturbations, while $\ell_\infty$ threats show clearer discrimination at global image-level features due to their uniform perturbations across the image.
We thus propose \textbf{threat-distinguishing dual-scale gating} that combines both scales to provide richer discriminative signals to gating networks.
However, even with richer input signals, the gating network may still converge to threat-agnostic expert assignments without explicit routing supervision.
We thus further introduce \textbf{threat-guided gating diversification}, which supervises gating networks to enforce diverse routing patterns across threats.

Building on these insights, we propose Robust Mixture of Low-Rank Experts (\sys) consisting of: (i) low-rank experts (Sec.~\ref{ssec:low_rank_experts}) for learning threat-specific model pathways, (ii) threat-distinguishing dual-scale gating (Sec.~\ref{ssec:bilevel_gating}) for discriminative gating signals, and (iii) threat-guided gating diversification (Sec.~\ref{ssec:threat_guided}) for diverse expert utilization across threats.
These components effectively route each threat to distinct expert combinations (Fig.~\ref{fig:teaser_concept}), with each expert learning distinct threat-specific features (Fig.~\ref{fig:tsne_ours}), enabling threat-adaptive model pathways that mitigate cross-threat trade-offs.

We conduct extensive experiments on CIFAR-10, ImageNet-100 and ImageNet-1K, demonstrating that \sys outperforms existing state-of-the-art MAT methods in union robustness and natural accuracy (Sec.~\ref{ssec:main_results}).
Diverse representations captured with multiple model pathways also improve robustness against threats unseen during training (Sec.~\ref{ssec:main_results}).
Ablation studies (Sec.~\ref{ssec:ablation}) validate the importance of each component of \sys, and analysis (Sec.~\ref{ssec:analysis_gating}) shows that our framework enables threat-specific representations for improved performance.

In summary, our contributions are as follows:
\begin{itemize}
    \item We identify threat-agnostic routing problem when applying MoE to MAT, where the gating network learns similar expert routing for different threats.
    \item To this end, we introduce \sys, a novel mixture of low-rank experts framework for adversarial training on multiple threat types via threat-distinguishing dual-scale gating and threat-guided gating diversification.
    \item Extensive experiments demonstrate that our approach outperforms existing state-of-the-art MAT in multi-threat robustness across multiple benchmarks.
\end{itemize}
\section{Related Works}
\label{sec:related_works}
\subsection{Multi-Perturbation Adversarial Training (MAT)}
Adversarial training, which minimizes the worst-case loss over adversarial examples~\cite{fgsm, pgd}, has become the de facto adversarial defense.
Subsequent work improved trade-off between robustness and natural accuracy~\cite{trades, mart}, enhanced training efficiency~\cite{free, fast}, and improved scalability to large-scale datasets~\cite{intriguing, revisiting, advxl}.
Despite these advances, most methods optimize for a single adversarial threat (\eg, $\ell_\infty$-bounded), limiting robustness against different threats.

To improve robustness across diverse adversarial threats, early work~\cite{max} proposed training a model on multiple $\ell_p$ threats.
Subsequent works improved robustness or training efficiency via worst-case descent directions~\cite{msd}, stochastic adversarial sampling~\cite{sat}, alternation between extreme norms~\cite{e-at}, parameter-space interpolation~\cite{soup}, logit-pairing with gradient projection~\cite{ramp}, or logit-space regularization~\cite{crt}. 
While effective, these methods learn robustness against distinct threats within a single model representation space, where optimizing for one threat can interfere with others~\cite{max, e-at, ramp} and degrade their robustness.
MORE~\cite{more} employs mixture-of-experts but relies on gating without explicit threat-aware guidance, which can face difficulty distinguishing between threats.
In contrast, our \sys learns diverse model pathways each tailored to different threats for mitigating cross-threat robustness trade-off.

\subsection{Mixture of Experts}
Mixture of Experts (MoE)~\cite{shazeer2017outrageously, jacobs1991adaptive} learns multiple experts and a gating network that routes each input to a subset of experts, enabling input-adaptive model specialization.
In Transformers, this is typically implemented by replacing feed-forward networks (FFNs) with multiple independent FFN experts~\cite{moe-survey, shazeer2017outrageously, switch-transformers}.
Different variants have enhanced its routing stability~\cite{switch-transformers, deepseekmoe}, scalability~\cite{gshard}, and expert selection~\cite{softmoe, zhou2022mixture}.
MoE has been applied across multi-task learning~\cite{ma2018modeling, adamv, mtl-lora}, continual learning~\cite{li2406theory, ddas}, and large-scale vision models~\cite{riquelme2021scaling}. 
More recently, MoE has been applied with parameter-efficient adapters~\cite{mole, loramoe, mola, rocket}.
While parallel lines of works explored adversarial robustness of MoE architectures~\cite{zhang2025optimizing, pavlitska2024towards} or used MoE for improving robustness against a single threat~\cite{puigcerver2022adversarial, zhang2023robust, meymani2025defending, pavlitska2025robust}, our work addresses an orthogonal problem of resolving cross-threat robustness trade-offs when training on multiple distinct threat types.
To this end, we tackle the redundant features and threat-agnostic routing issues of conventional MoE through low-rank experts, dual-scale gating, and gating diversification.

\section{Preliminary}
\label{sec:prelim}
We consider a classification model $f_\theta: \mathcal{X} \rightarrow \mathcal{Y}$ parameterized by $\theta$, where $\mathcal{X}$ and $\mathcal{Y}$ denote the input and label spaces, and $(x, y) \sim \mathcal{D}$ is a sample from data distribution $\mathcal{D}$.
Given adversarial threats $\{\mathcal{A}_1, \mathcal{A}_2, \mathcal{A}_\infty\}$ with perturbation budgets $\{\epsilon_1, \epsilon_2, \epsilon_\infty\}$, multi-perturbation adversarial training (MAT) aims to solve:
\begin{equation}
    \min_{\theta} \mathbb{E}_{(x,y) \sim \mathcal{D}} \, \phi(\mathcal{L}_1, \mathcal{L}_2, \mathcal{L}_\infty),
    \label{eq:multi-at-base}
\end{equation}
where $\mathcal{L}_p = \max_{\delta_p \in \mathcal{A}_p(x, \epsilon_p)} \mathcal{L}(f_\theta(x + \delta_p), y)$ is adversarial loss~\cite{pgd} under threat $\mathcal{A}_p$ for $p \in \{1, 2, \infty\}$~\cite{max, e-at, ramp}, and $\phi: \mathbb{R}^3 \rightarrow \mathbb{R}$ aggregates losses across threats.
\textbf{MAX}~\cite{max} implements aggregation as $\phi_{\text{MAX}}(\mathcal{L}_1, \mathcal{L}_2, \mathcal{L}_\infty) = \max \{\mathcal{L}_1, \mathcal{L}_2, \mathcal{L}_\infty\}$ to optimize on the worst-case threat at each iteration.
\textbf{RANDOM}~\cite{sat} implements aggregation as $\phi_{\text{RANDOM}}(\mathcal{L}_1, \mathcal{L}_2, \mathcal{L}_\infty) = \mathcal{L}_p$, where a threat $p \sim \text{Cat}(\frac{1}{3}, \frac{1}{3}, \frac{1}{3})$ is sampled stochastically.

\section{\sys}
\label{sec:methods}
\subsection{Limitations of Conventional MoE}
\label{ssec:motivations}

\begin{figure}[t]
    \centering
    
    \includegraphics[width=0.5\linewidth]{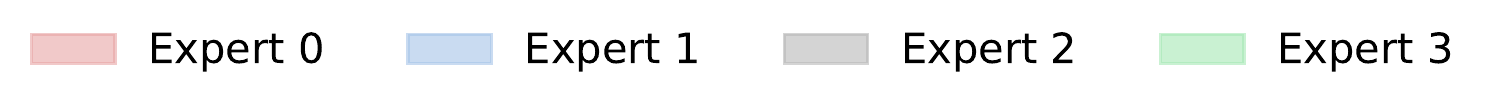} \\
    \medskip
    \vspace{-5pt}
    
    \begin{subfigure}{0.32\linewidth}
        \centering
        \includegraphics[width=\linewidth, trim=5 5 5 5, clip]{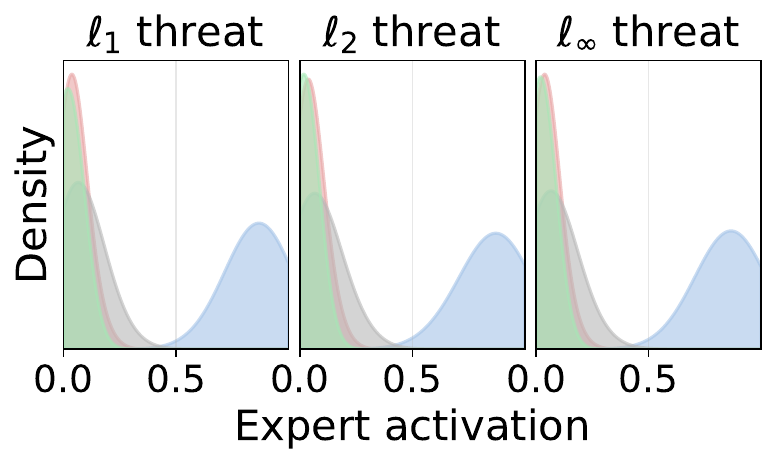}
        \caption{Na\"ive approach}
        \label{fig:gating_diversity_naive}
    \end{subfigure}
    \hfill
    \begin{subfigure}{0.32\linewidth}
        \centering
        \includegraphics[width=\linewidth, trim=5 5 5 5, clip]{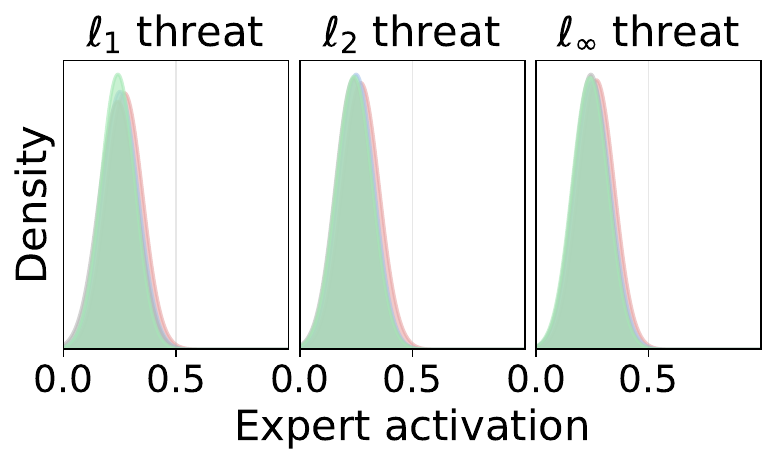}
        \caption{Load balancing~\cite{shazeer2017outrageously}}
        \label{fig:gating_diversity_loadbal}
    \end{subfigure}
    \hfill
    \begin{subfigure}{0.32\linewidth}
        \centering
        \includegraphics[width=\linewidth, trim=5 5 5 5, clip]{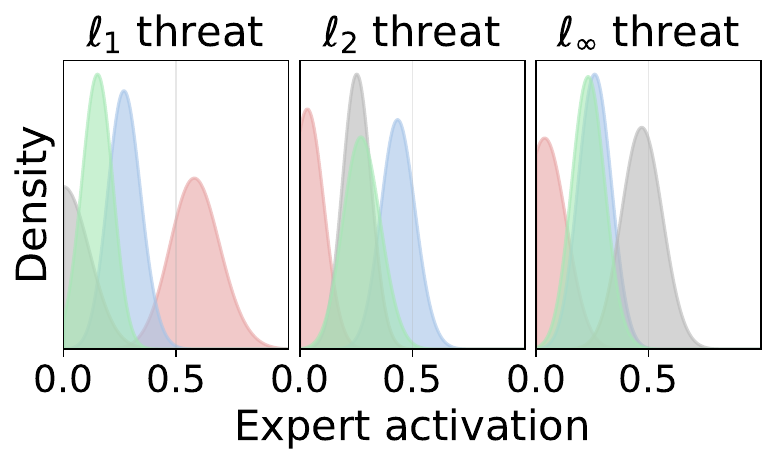}
        \caption{Ours (\sys)}
        \label{fig:gating_diversity_ours}
    \end{subfigure}
    
    \caption{
    Gaussian KDE analysis on expert activations across different threat types on CIFAR-10.
    (a) Na\"ive MoE suffers from routing collapse (Expert 1 dominates) and threat-agnostic routing (similar routing across threats). 
    (b) Load balancing~\cite{shazeer2017outrageously} resolves routing collapse but exhibits threat-agnostic routing.
    (c) Our \sys addresses threat-agnostic routing issue and learns diverse, threat-specific expert combinations.
    }
    \label{fig:gating_diversity}
    \vspace{-5pt}
\end{figure}

We employ mixture of experts (MoE) to construct multiple threat-specific model pathways for improved multi-threat robustness.
However, unlike prior MoE applications with clearly distinguishable inputs (\eg, different tasks or languages)~\cite{sparse-moe, moe-lpr, adamv, mod, m3vit}, adversarial threats share the identical semantic content of the underlying image being perturbed, which creates a strong inductive bias~\cite{not-bugs, humans-decipher}.
Without an architectural component to capture these common characteristics~\cite{mtl-lora}, experts are prone to capturing redundant similar representations as shown in Fig.~\ref{fig:teaser_tsne}, thus preventing the construction of threat-specific model pathways.

These threat-common features further induce \emph{threat-agnostic routing} problem, where the gating network receives insufficient discriminative signal and routes different threats to similar expert combinations, failing to mitigate cross-threat trade-offs.
To demonstrate this issue, we analyze the gating weight distributions across threats in Fig.~\ref{fig:gating_diversity}.
We train ViT-B~\cite{vit} on CIFAR-10~\cite{cifar10} with 4 FFN experts~\cite{moe-survey, shazeer2017outrageously} for each Transformer block using RANDOM~\cite{sat} (full details in Sec.~\ref{sec:suppl_experiment_setup} of supplementary) and visualize token-averaged gating weights from the final layer over 10K test samples.

We first consider a na\"ive approach that trains the MoE with adversarial loss only (Eq.~\ref{eq:multi-at-base}).
As shown in Fig.~\ref{fig:gating_diversity_naive}, this approach faces (i) \emph{routing collapse}, where only a few experts (Expert 1) dominate, and (ii) \emph{threat-agnostic routing}, where all threats exhibit nearly identical gating distributions.
To address routing collapse, we apply load balancing loss~\cite{shazeer2017outrageously, mole} widely used in conventional MoE~\cite{shazeer2017outrageously, qwen3, deepseekmoe} to encourage uniform expert utilization, with its loss coefficient set to $0.5$ following existing protocol~\cite{mole}.
As shown in Fig.~\ref{fig:gating_diversity_loadbal}, while load balancing resolves routing collapse, it still suffers from threat-agnostic routing, where the gating distributions remain similar across threats.
Evaluation across different loss coefficients in Fig.~\ref{fig:gating_diversity_lam} also consistently exhibit threat-agnostic routing problem.
This reveals that existing MoE techniques alone are insufficient, motivating our framework to learn distinct, threat-aware expert combinations (Fig.~\ref{fig:gating_diversity_ours}).

\begin{figure*}[t]
    \centering
    \includegraphics[width=\linewidth]{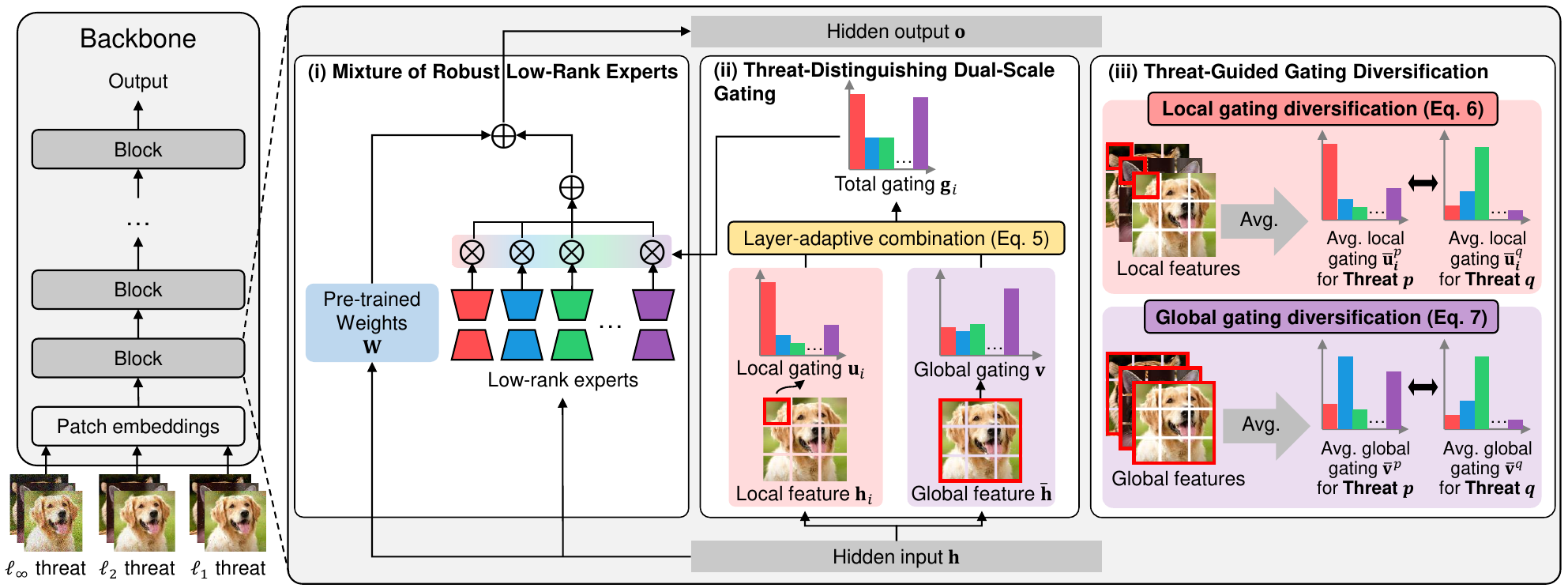}
    \caption{
    Overview of \sys.
    (i) We train a mixture of robust low-rank experts, where a shared backbone captures threat-common features while low-rank experts focus on threat-specific information.
    To address threat-agnostic routing, where the gating assigns similar expert combinations across threats, we propose (ii) threat-distinguishing dual-scale gating, which leverages local patch-level and global image-level features to better distinguish between threats, and (iii) threat-guided gating diversification, which ensures different threats are routed to distinct expert combinations.
    }
    \label{fig:framework}
\end{figure*}

In this work, we propose \sys (Robust Mixture of Low-Rank Experts), illustrated in Fig.~\ref{fig:framework}, consisting of three components: (i) low-rank experts (Sec.~\ref{ssec:low_rank_experts}) to capture threat-specific features, (ii) threat-distinguishing dual-scale gating (Sec.~\ref{ssec:bilevel_gating}) to provide gating with threat-discriminative signals, and (iii) threat-guided gating diversification (Sec.~\ref{ssec:threat_guided}) to enforce routing diversity across threats.

\subsection{Mixture of Robust Low-Rank Experts}
\label{ssec:low_rank_experts}
To prevent experts from redundantly learning shared information common across threats, we implement each expert as a low-rank additive update to the backbone weights.
We allow the shared backbone to capture threat-common features, while the low-rank experts focus on capturing threat-specific information.

Inspired by its recent success, we implement these low-rank experts as a LoRA~\cite{lora} module.
Each expert $k \in \{1, \dots, K\}$ learns an additive adjustment to the backbone layer with weight matrix $\mathbf{W} \in \mathbb{R}^{d_\text{out} \times d_\text{in}}$ via low-rank matrices $\mathbf{B}_k \in \mathbb{R}^{d_\text{out} \times r}$ and $\mathbf{A}_k \in \mathbb{R}^{r \times d_\text{in}}$, where $r \ll d_\text{out}, d_\text{in}$.
Given input feature $\mathbf{h}_i \in \mathbb{R}^{d_\text{in}}$ at token $i$, the output $\mathbf{o}_i \in \mathbb{R}^{d_\text{out}}$ is computed as:
\begin{equation}
    \mathbf{o}_i = \mathbf{W}\mathbf{h}_i + \sum^{K}_{k=1}{g_{i,k} \cdot \mathbf{B}_k \mathbf{A}_k \mathbf{h}_i}.
    \label{eq:lora-moe}
\end{equation}
Gating weight $\mathbf{g}_i = [g_{i,1}, \ldots, g_{i,K}]^\top \in \mathbb{R}^{K}$ is predicted for each token $i$ (\ie, patch) in an image by a gating network such that $g_{i,k} \in [0, 1]$ and $\sum^K_{k=1}g_{i,k} = 1$.

\subsection{Threat-Distinguishing Dual-Scale Gating}
\label{ssec:bilevel_gating}
To address threat-agnostic routing, we propose dual-scale gating that leverages both local patch-level and global image-level features to distinguish between threats.
This is based on the intuition that adversarial threats exhibit discriminative patterns at different feature levels.
For instance, $\ell_1$ threats create sparse perturbations on specific pixels, best captured by patch-level features, while $\ell_\infty$ perturbations are spatially uniform and are more evident in image-level features.

\begin{figure}[t]
    \centering 
    
    \begin{subfigure}[b]{0.35\textwidth}
        \centering
        \includegraphics[width=\linewidth, trim=10 10 10 10, clip]{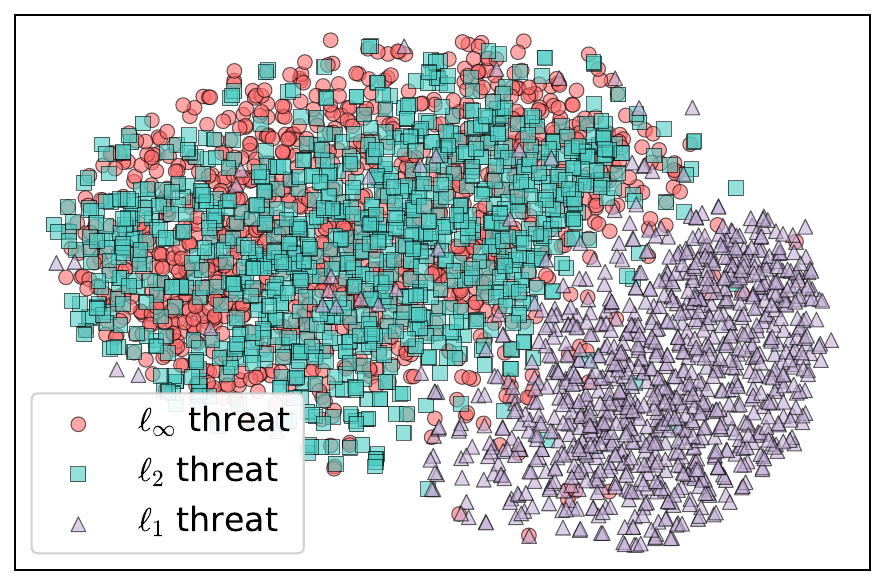}
        \caption{Patch-level features}
        \label{fig:tsne_local}
    \end{subfigure}
    \hspace{0.05\textwidth} 
    \begin{subfigure}[b]{0.35\textwidth}
        \centering
        \includegraphics[width=\linewidth, trim=10 10 10 10, clip]{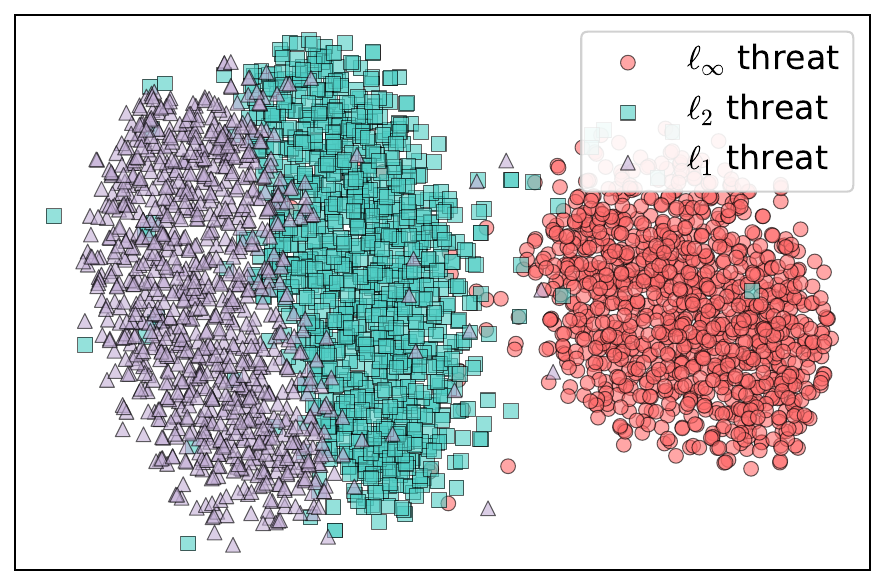}
        \caption{Image-level features}
        \label{fig:tsne_global}
    \end{subfigure}
    
    \caption{
    t-SNE visualization of threat separability at different feature levels.
    (a) Local patch-level features better separate $\ell_1$ threats (silhouette: 0.394) but show overlap for $\ell_\infty$ and $\ell_2$.
    (b) Global image-level features show the complementary pattern, better distinguishing $\ell_\infty$ (silhouette: 0.611) over $\ell_1$ and $\ell_2$.
    }
    \label{fig:tsne_dual}
\end{figure}

To verify this, we visualize feature separability using t-SNE~\cite{tsne} (Fig.~\ref{fig:tsne_dual}) from ViT-B~\cite{vit} trained with RANDOM strategy~\cite{sat} on CIFAR-10.
For patch-level features (Fig.~\ref{fig:tsne_local}), we compute separability for all $T$ token positions through silhouette coefficient score~\cite{silhouette}.
For image-level features (Fig.~\ref{fig:tsne_global}), we average features across all patches and compute separability score.
Our analysis reveals an interesting observation that local features show substantially higher separability for $\ell_1$ threats (mean silhouette score: $0.394 \pm 0.155$) compared to $\ell_\infty$ ($0.125 \pm 0.077$) and $\ell_2$ ($0.078 \pm 0.053$).
Conversely, global image-level features exhibit the opposite trend ($\ell_1$: $0.303$ , $\ell_2$: $0.281$, $\ell_\infty$: $0.611$), indicating that $\ell_\infty$ threats are more distinguishable compared to $\ell_1$ and $\ell_2$ threats.
Analysis across layers and patches in Fig.~\ref{fig:tsne_all_patch} verifies the same phenomenon.

Based on these observations, we design a dual-scale gating comprised of local gating and global gating, each capturing patch-wise and image-level perturbation patterns, respectively.
Given the hidden state $\mathbf{h}_i \in \mathbb{R}^{d_\text{in}}$ at token $i$, the local gating network $\text{MLP}_{\text{local}}$ predicts patch-wise gating weight $\mathbf{u}_{i}$ as follows:
\begin{equation}
    \mathbf{u}_{i} = \text{softmax}\left(\text{MLP}_{\text{local}}(\mathbf{h}_i)\right) \in \mathbb{R}^{K}.
    \label{eq:local_gate}
\end{equation}
For image-level information, we average hidden states across all $T$ tokens to obtain $\bar{\mathbf{h}} = \frac{1}{T} \sum^{T}_{i=1} \mathbf{h}_i$ and pass it through the global gating network $\text{MLP}_{\text{global}}$:
\begin{equation}
    \mathbf{v} = \text{softmax}\left(\text{MLP}_{\text{global}}(\bar{\mathbf{h}})\right) \in \mathbb{R}^{K}.
    \label{eq:global_gate}
\end{equation}
This outputs a single gating weight vector $\mathbf{v}$ for an image.

\vspace{0.5ex}\noindent
\textbf{Layer-adaptive combination.}
Early Transformer layers encode token-level information while deeper layers capture global semantic-level features~\cite{hmora, geva2020transformer}, creating a layer-wise bias where global features are under-represented in early layers and local features in deeper layers.
Since both features are crucial for threat discrimination (Fig.~\ref{fig:tsne_dual}), we compensate for this bias as:
\begin{equation}
    \mathbf{g}_i = (1 - \beta(l)) \cdot \mathbf{u}_{i} + \beta(l) \cdot \mathbf{v},
    \label{eq:layer_adaptive}
\end{equation}
where $\mathbf{g}_i = [ g_{1,i}, \ldots, g_{K,i} ]^T$ is the final gating weight for token $i$, and $g_{k,i}$ denotes the weight for expert $k$.
The layer-adaptive coefficient $\beta(l) = \sigma\left(s \cdot \left( 1 - \frac{l}{L} \right) - b\right) \in [0, 1]$ emphasizes global gating in early layers and local gating in deeper layers, with $\sigma(\cdot)$ the sigmoid function, $L$ the total number of layers, and $s$, $b$ controlling transition across layers~\cite{hmora}.
Since the structural property of Transformer layer hierarchy is well-established~\cite{hmora, geva2020transformer}, we encode it directly as a fixed prior, which empirically outperforms the learnable variant (Table~\ref{tab:ablation} in Sec.~\ref{ssec:ablation}).

\subsection{Threat-Guided Gating Diversification}
\label{ssec:threat_guided}
While our dual-scale gating captures discriminative cues for each threat, it still requires explicit supervision to learn diverse, threat-specific expert combinations.
To this end, we introduce a regularization loss that encourages the gating network to assign distinct expert combinations across threats, effectively constructing separate model pathways for each threat type.

Let $\mathcal{B}_p$ denote adversarial examples under threat $\mathcal{A}_p$ for $p \in \{1, 2, \infty\}$. 
For each threat $p$ and token $i$, we compute the average local gating $\bar{\mathbf{u}}_{i}^p$ across $\mathcal{B}_p$.
We then maximize pairwise Euclidean distances between threat-specific gating patterns, aggregating across all $T$ token positions:
\begin{equation}
    \mathcal{L}_{\text{div-local}} = -\frac{1}{T} \sum_{p \neq q} \sum_{i=1}^{T} 
    \left\| \bar{\mathbf{u}}_{i}^p - \bar{\mathbf{u}}_{i}^q \right\|_2^2.
    \label{eq:local_div}
\end{equation}
Since global gating weights are $K$-dimensional (number of experts), we project them to a higher $K'$ dimension via a linear layer for finer-grained threat discrimination. 
The average projected gating $\bar{\mathbf{v}}^{p}$ for threat $p$ yields the following loss:
\begin{equation}
    \mathcal{L}_{\text{div-global}} = -\sum_{p \neq q} \left\| \bar{\mathbf{v}}^{p} - \bar{\mathbf{v}}^{q} \right\|_2^2.
    \label{eq:global_div}
\end{equation}
This projection is used for computing diversification loss and does not affect routing in Eq.~\ref{eq:global_gate}.
In Table~\ref{tab:distance_choice}, we explore other options for measuring the distance between gating weights, out of which our design leads to the best performance. 

We aggregate both objectives for each layer $l$ across all $L$ layers to encourage gating networks to learn distinct, threat-specific expert routing:
\begin{equation}
\label{eq:total_div}
    \mathcal{L}_{\text{div}} = \frac{1}{L} \sum_{l=1}^{L} \left( \mathcal{L}^{(l)}_{\text{div-local}} + \mathcal{L}^{(l)}_{\text{div-global}} \right).
\end{equation}

The diverse model pathways learned by the experts provide broader coverage of the perturbation space, improving robustness against unseen threats compared to existing MAT methods that rely on a single model pathway.
Furthermore, since threat labels are only used during training, the learned gating predicts the optimal expert combination at inference without requiring knowledge of the input threat type, allowing our method to generalize to unseen threats. 
Notably, this generalization extends beyond $\ell_p$ threats and improves robustness also against various non-$\ell_p$ threats (Table~\ref{tab:unseen_attacks_ext}, Sec.~\ref{ssec:main_results}).

\subsection{Training Objective}
\label{ssec:training_objective}
Thanks to its modular design, \sys can be integrated with existing MAT methods.
In this work, we apply it to MAX~\cite{max} and RANDOM~\cite{sat}.
The training objective combines the base loss (Eq.~\ref{eq:multi-at-base}) with our diversification regularization:
\begin{equation}
    \min_{\theta} \mathbb{E}_{(x,y) \sim \mathcal{D}} \left[ \phi\left(\mathcal{L}_1, \mathcal{L}_2, \mathcal{L}_\infty\right) \right] + \lambda \mathcal{L}_{\text{div}},
\end{equation}
where $\phi$ denotes aggregation function $\phi_\text{MAX}$ or $\phi_\text{RANDOM}$, and $\lambda$ controls strength of $\mathcal{L}_{\text{div}}$.
$\theta$ denotes parameters for backbone, experts, and gating networks.

While this section focuses on $\ell_p$-MAT, modular design of our \sys makes it widely applicable to other adversarial training methods.
In Sec.~\ref{ssec:main_results}, we demonstrate this by applying \sys to non-$\ell_p$ perceptual adversarial training~\cite{pat, vr} and show that \sys further improves their robustness against unseen threats.

\section{Experiments}
\label{sec:experiments}
\subsection{Experimental Setup}
\vspace{0.5ex}\noindent
\textbf{Datasets, models, and baselines.}
We evaluate on CIFAR-10~\cite{cifar10}, ImageNet-100~\cite{imagenet100}, and ImageNet-1K~\cite{imagenet} using standard ImageNet-1K pre-trained~\cite{imagenet, timm} ViT-B~\cite{vit}, DeiT-B~\cite{deit}, and Swin-B~\cite{swin}.
For MAT, we compare with RANDOM~\cite{sat}, AVG~\cite{max}, MAX~\cite{max}, MSD~\cite{msd}, MORE~\cite{more}, E-AT~\cite{e-at}, and RAMP~\cite{ramp}.
For evaluation against unseen threats, we follow OODRobustBench~\cite{oodrobustbench} and also compare with PAT~\cite{pat} and VR~\cite{vr}, two representative non-$\ell_p$ adversarial training methods trained on AlexNet~\cite{alexnet}-based and self-model perceptual threats. 
For all baselines, we use default hyperparameters from their original manuscripts.

\vspace{0.5ex}\noindent
\textbf{Threat setup.}
Following existing protocols~\cite{max, e-at, ramp}, we train and evaluate under PGD~\cite{pgd} and APGD~\cite{apgd} with perturbation budgets $(\epsilon_1, \epsilon_2, \epsilon_\infty)$ of $(12, 0.5, \frac{8}{255})$ for CIFAR-10 and $(255, 2.0, \frac{4}{255})$ for ImageNet-100, 1K.
For PGD~\cite{max, msd, sat}, we use steps $(n_1, n_2, n_\infty)$ of $(20, 20, 10)$ for CIFAR-10 and $(40, 20, 10)$ for ImageNet-100, 1K.
For APGD~\cite{e-at, ramp}, we use steps $n = 15$ for both CIFAR-10 and ImageNet-100, 1K.
For robust fine-tuning, following E-AT~\cite{e-at} and RAMP~\cite{ramp}, we use APGD with $10$ steps on CIFAR-10 and $5$ ($\ell_\infty$/$\ell_2$) and $15$ ($\ell_1$) steps on ImageNet-1K.
For evaluation, we use AutoAttack under $\ell_1$, $\ell_2$, and $\ell_\infty$ norms.
We evaluate natural accuracy on clean images, per-threat robustness, their average, and worst-case union robustness across all threats~\cite{ramp}.

\begin{table*}[t]
    \centering
    \renewcommand{\arraystretch}{1.15}
    \setlength{\tabcolsep}{6pt}
    \caption{
    Comparison with state-of-the-art methods on CIFAR-10 and ImageNet-100 under AutoAttack.
    We measure natural accuracy on clean images, single-threat robustness under $\ell_1$, $\ell_2$, and $\ell_\infty$, their average, and worst-case union robustness.
    Best results are marked in \textbf{bold}.
    }
    \resizebox{\textwidth}{!}{
    \begin{tabular}{c|c|cccccc|cccccc}
        \specialrule{2pt}{\aboverulesep}{\belowrulesep}
        \multicolumn{2}{c|}{\multirow{2}{*}{Methods}}
        & \multicolumn{6}{c|}{\textbf{\textit{CIFAR-10}} }
        & \multicolumn{6}{c}{\textbf{\textit{ImageNet-100}}} \\
        \cmidrule{3-14}
        \multicolumn{2}{c|}{} & Nat & $\ell_1$ & $\ell_2$ & $\ell_\infty$ & Avg & Union
            & Nat & $\ell_1$ & $\ell_2$ & $\ell_\infty$ & Avg & Union \\
        \midrule
        \multirow{7}{*}{PGD}
        & RANDOM
        & 86.7 & 52.3 & 68.8 & 40.3 & 53.8 & 38.6
        & 84.4 & 43.8 & 67.6 & 45.2 & 52.2 & 39.4 \\
        & AVG
        & 86.3 & 52.2 & 67.5 & 38.9 & 52.9 & 37.9
        & 83.6 & 44.3 & 66.4 & 44.8 & 51.8 & 39.3 \\
        & MAX
        & 85.1 & 46.6 & 66.8 & 40.1 & 51.2 & 38.5
        & 79.9 & 42.2 & 63.2 & 47.5 & 51.0 & 40.5 \\
        & MSD
        & 84.7 & 43.9 & 67.3 & 40.5 & 50.6 & 38.3
        & 80.2 & 42.6 & 64.3 & 45.8 & 50.9 & 40.2 \\
        & MORE
        & 82.6 & 44.3 & 42.4 & 33.7 & 40.1 & 31.1
        & 75.6 & 34.5 & 41.2 & 42.0 & 39.2 & 32.8 \\
        \rowcolor{ourcolor} & RoME+RANDOM (Ours)
        & \textbf{88.1} & \textbf{56.4} & \textbf{69.7} & 42.1 & \textbf{56.1} & 41.1
        & \textbf{85.2} & \textbf{46.9} & \textbf{68.3} & 47.8 & \textbf{54.3} & 42.9 \\
        \rowcolor{ourcolor} & RoME+MAX (Ours)
        & 85.4 & 48.2 & 68.2 & \textbf{43.1} & 53.2 & \textbf{42.3}
        & 80.6 & 45.6 & 66.0 & \textbf{50.9} & 54.2 & \textbf{43.7} \\
        \midrule \midrule
        \multirow{6}{*}{APGD}
        & RANDOM
        & 86.5 & 50.1 & 69.3 & 37.5 & 52.3 & 37.2
        & 82.0 & 38.4 & 66.0 & 46.2 & 50.2 & 37.1 \\
        & MAX
        & 81.7 & 45.5 & 65.3 & 42.6 & 51.1 & 41.4
        & 79.5 & 41.2 & 63.6 & 44.8 & 49.9 & 39.8 \\
        & E-AT
        & 83.2 & 50.3 & 68.9 & 40.6 & 53.3 & 39.7
        & \textbf{83.4} & 40.6 & 64.2 & 45.4 & 50.1 & 39.2 \\
        & RAMP
        & 82.2 & 47.1 & 63.5 & 43.4 & 51.3 & 42.5
        & 82.2 & 42.6 & 63.4 & 45.1 & 50.4 & 41.8 \\
        \rowcolor{ourcolor} & RoME+RANDOM (Ours)
        & \textbf{87.4} & \textbf{54.7} & \textbf{71.6} & 40.0 & \textbf{55.4} & 39.4
        & \textbf{83.4} & 42.1 & \textbf{67.8} & \textbf{48.6} & \textbf{52.8} & 40.5 \\
        \rowcolor{ourcolor} & RoME+MAX (Ours)
        & 82.5 & 48.9 & 67.3 & \textbf{44.2} & 53.5 & \textbf{43.7}
        & 80.2 & \textbf{44.2} & 64.3 & 47.9 & 52.1 & \textbf{42.8} \\
        \specialrule{2pt}{\aboverulesep}{\belowrulesep}
    \end{tabular}}
    \label{tab:sota}
\end{table*}

\vspace{0.5ex}\noindent
\textbf{Implementation details.}
For fair comparison, we follow standard Transformer training recipe~\cite{vit, deit} for all baselines and our method, using AdamW~\cite{adamw} with initial learning rates of $\texttt{1e-3}$ for CIFAR-10 and $\texttt{1e-4}$ for ImageNet-100, 1K, with 
warmup and linear decay scheduling (full details in Sec.~\ref{sec:suppl_experiment_setup}).
We train for 20 epochs on CIFAR-10 and 5 epochs on ImageNet-100, 1K.
For robust fine-tuning, we fine-tune $\ell_\infty$-robust models for $3$ epochs on CIFAR-10 and $1$ epoch on ImageNet-1K.
We use $K = 4$ experts of rank $16$, apply them to $QKV$ and $O$ projection layers in all Transformer blocks, and set $\lambda = 0.1$, $s = 4$, $b = 2$, and $K' = 100$ for global gating projection.
We also apply our method to CNNs (WideResNet~\cite{wideresnet}), where low-rank experts on convolutional layers are implemented using the official LoRA implementation~\footnote{\url{https://github.com/microsoft/LoRA/blob/main/loralib/layers.py}}, and local and global features are obtained per spatial location of the convolutional feature map and by global average pooling over all spatial locations, respectively.

\begin{table*}[t]
    \centering
    \renewcommand{\arraystretch}{1.15}
    \setlength{\tabcolsep}{5pt}
    \caption{
    Comparison with prior methods on CIFAR-10 under unseen common corruptions and adversarial threats. 
    Percep. $\ell_p$ is the average of Fog~\cite{pat}, Snow~\cite{pat}, Gabor~\cite{pat}, Elastic~\cite{pat}, and $\ell_\infty$-Jpeg~\cite{pat} robustness.
    Best results are marked in \textbf{bold}.
    }
    \resizebox{\textwidth}{!}{
    \begin{tabular}{c|c|ccccccccc}
        \specialrule{2pt}{\aboverulesep}{\belowrulesep}
        \multicolumn{2}{c|}{Methods} & Com. Corr. & $\ell_0$ & Percep. $\ell_p$ & PPGD & LPA & Adv. Patch & StAdv & ReColorAdv & GMA \\
        \midrule
        \multirow{7}{*}{PGD} & 
        RANDOM
        & 79.0 & 31.9 & 53.2 & 43.7 & 32.2 & 47.9 & 31.1 & 70.4 & 44.0 \\
        & AVG
        & 78.4 & 26.7 & 54.5 & 42.9 & 31.0 & 48.6 & 29.2 & 68.8 & 43.7 \\
        & MAX
        & 76.8 & 28.3 & 51.3 & 43.2	& 33.1 & 43.9	& 34.4	& 70.0 & 45.4 \\
        & MSD
        & 76.5 & 28.0 & 50.4 & 42.4	& 33.3 & 40.8	& 34.9	& 70.8 & 45.6 \\
        & MORE
        & 74.9 & 23.8 & 50.7 & 42.7 & 31.5 & 45.2 & 32.1 & 70.0 & 43.6 \\
        \rowcolor{ourcolor} & RoME+RANDOM (Ours)
        & \textbf{80.2} & 38.3 & \textbf{57.8} & 51.2	& 35.4 & \textbf{48.8}	& 36.8	& 73.5 & 45.6 \\
        \rowcolor{ourcolor} & RoME+MAX (Ours)
        & 76.8 & \textbf{39.4} & 56.2 & \textbf{52.2}	& \textbf{40.7} & 47.4	& \textbf{44.0}	& \textbf{75.4} & \textbf{47.5} \\
        \midrule \midrule
        \multirow{6}{*}{APGD} & 
        RANDOM
        & 78.5 & 31.6 & 54.5 & 49.5	& 45.5 & 49.1	& 47.0	& 77.8 & 40.0 \\
        & MAX
        & 74.2 & 33.4 & 51.6 & 50.5	& 49.2 & 47.2	& 54.6	& 76.6 & 44.9 \\
        & E-AT
        & 74.3 & 34.1 & 49.7 & 49.7 & 47.9 & 48.0 & 48.5 & 75.5 & 42.8 \\
        & RAMP
        & 74.9 & 33.1 & 51.6 & 51.0 & 50.0 & 49.4 & 49.9 & 76.5 & 45.1 \\
        \rowcolor{ourcolor} & RoME+RANDOM (Ours)
        & \textbf{79.6} & \textbf{38.8} & \textbf{58.9} & 50.9	& 47.0 & \textbf{50.4}	& 49.2	& \textbf{78.2} & 43.3 \\
        \rowcolor{ourcolor} & RoME+MAX (Ours)
        & 74.5 & 37.8 & 53.0 & \textbf{51.5}	& \textbf{51.6} & 49.2	& \textbf{56.0}	& 78.0 & \textbf{45.6} \\
        \midrule \midrule
        &
        PAT
        & 77.6 & 16.0 & 47.1 & 55.5	& 51.5 & 41.2	& 59.7	& 76.4 & 49.0 \\
        & PAT+VR
        & 77.7 & 9.9 & 47.9 & 56.6	& 51.9 & 39.3	& 58.4	& 77.0 & 50.1 \\
        \rowcolor{ourcolor} & RoME+PAT (Ours)
        & 78.1 & \textbf{19.8} & 48.6 & 57.0	& 54.9 & 45.2	& 63.3	& 77.5 & 49.8 \\
        \rowcolor{ourcolor} \multirow{-4}{*}{Non-$\ell_p$} & RoME+PAT+VR (Ours)
        & \textbf{78.3} & 17.9 & \textbf{49.1} & \textbf{59.3}	& \textbf{57.8} & \textbf{45.7}	& \textbf{65.2}	& \textbf{80.1} & \textbf{52.6} \\
        \specialrule{2pt}{\aboverulesep}{\belowrulesep}
    \end{tabular}}
    \label{tab:unseen_attacks_ext}
\end{table*}

\subsection{Multi-Perturbation Adversarial Robustness}
\label{ssec:main_results}
\vspace{0.5ex}\noindent
\textbf{Robustness against multiple $\ell_p$ perturbations.}
In Table~\ref{tab:sota}, we report natural accuracy and robustness against $\ell_p$ threats seen during training.
Our approach effectively mitigates cross-threat trade-off, achieving the best average and union robustness across all settings.
For example, our RoME+MAX improves union robustness of baseline MAX by 3.8\%p on CIFAR-10 under PGD and 3.0\%p on ImageNet-100 under APGD.
We further validate on ImageNet-1K (Table~\ref{tab:sota_imagenet}), where our method improves union robustness by 1.7\%p and 1.7\%p over RANDOM and MAX under PGD and by 1.3\%p and 1.6\%p under APGD, demonstrating its scalability.
It also consistently outperforms baselines on DeiT-B~\cite{deit} and Swin-B~\cite{swin} (Table~\ref{tab:sota_deit}).
Our approach also achieves the highest natural accuracy.
This is because prior MAT methods force all inputs into a single representation space, where the accuracy-robustness trade-off inherent in adversarial training~\cite{trades, mart} degrades performance on clean images.
In contrast, our \sys learns a unique model pathway for natural images differently from adversarial examples as shown in Fig.~\ref{fig:gating_activations} (Sec.~\ref{ssec:gating_activations}).

\vspace{0.5ex}\noindent
\textbf{Robustness against unseen attacks.}
Following previous benchmarks~\cite{e-at, ramp}, we evaluate on common corruptions~\cite{common-corruptions}, $\ell_0$ attack~\cite{sparsefool}, $\ell_p$ perceptual attacks~\cite{pat}, and AutoAttack~\cite{apgd}.
We also evaluate on non-$\ell_p$ threats, including perceptual PPGD~\cite{pat} and LPA~\cite{pat}, adversarial patch~\cite{patch}, spatial StAdv~\cite{stadv}, and semantic ReColorAdv~\cite{recoloradv} attacks.
We also design and evaluate against gating misrouting attack (GMA), a strong white-box adaptive attack with full access to the gating framework.
GMA optimizes $\mathcal{L}_{\text{GMA}} = \mathcal{L}_{\text{CE}}(f_\theta(x+\delta), y) - \frac{1}{L}\sum_{l=1}^{L}\frac{1}{T}\sum_{i=1}^{T}\mathcal{L}_{\text{CE}}(\mathbf{g}_i^{(l)}(x+\delta), k^*_i)$, where the first term maximizes the classification loss and the second forces misrouting of each token's gating weight $\mathbf{g}_i^{(l)}$ (Eq.~\ref{eq:layer_adaptive}) to the least likely expert $k^*_i = \arg\min_k \, g_{i,k}$.
For this attack, we use $\ell_\infty$-PGD with $\epsilon = \frac{8}{255}$ and $20$ steps.
Full attack configurations (budget and steps) are provided in Sec.~\ref{sec:suppl_experiment_setup}.

Table~\ref{tab:unseen_attacks_ext} shows that our approach outperforms existing MAT methods across all unseen threats.
Notably, while MORE~\cite{more} also leverages MoE, it shows lower robustness, highlighting that our approach of learning distinct model pathways improves generalization to unseen threats.
Our method also outperforms existing methods against non-$\ell_p$ threats (\eg, PPGD and LPA), demonstrating that the diverse representations captured across multiple model pathways generalize beyond the $\ell_p$ threats seen during training.
It maintains highest robustness even against adaptive attacks, verifying the robustness of our gating framework.

To demonstrate the modularity of our \sys (Sec.~\ref{ssec:training_objective}), we also compare with PAT~\cite{pat} and VR~\cite{vr}, two representative non-$\ell_p$ adversarial training methods designed for unseen threat robustness.
Different from $\ell_p$-MAT, we follow the protocol of OODRobustBench~\cite{oodrobustbench} and train model on AlexNet~\cite{alexnet}-based and self-model perceptual threats~\cite{pat}.
Applying our method on these methods consistently improves unseen threat robustness with gains of up to $6.8\%$p against StAdv, demonstrating that \sys is broadly applicable beyond $\ell_p$-MAT.

\begin{table} [t]
    \centering
    \caption{
    Comparison with existing methods on robust fine-tuning scenario across different model architectures pre-trained on $\ell_\infty$-adversarial training with different datasets.
    Best results are marked in \textbf{bold}.
    }
    \label{tab:finetune}
    \setlength{\tabcolsep}{10pt}   
    \resizebox{\columnwidth}{!}{
    \begin{tabular}{l cc cc cc cc}
        \toprule
        & \multicolumn{2}{c}{\makecell{ViT-B\\(CIFAR-10)}} & \multicolumn{2}{c}{\makecell{WRN-28-10\\(CIFAR-10)}} & \multicolumn{2}{c}{\makecell{WRN-94-16\\(CIFAR-10)}} & \multicolumn{2}{c}{\makecell{XCiT-S\\(ImageNet-1K)}} \\
        \cmidrule(lr){2-3} \cmidrule(lr){4-5} \cmidrule(lr){6-7} \cmidrule(lr){8-9}
        Method & Nat & Union & Nat & Union & Nat & Union & Nat & Union \\
        \midrule
        E-AT        & 83.6 & 41.7 & 89.4 & 50.3 & 92.8 & 44.7 & 65.0 & 25.9 \\
        \quad +RoME & \textbf{84.7} & \textbf{43.5} & \textbf{91.3} & \textbf{51.8} & \textbf{93.3} & \textbf{48.4} & \textbf{66.1} & \textbf{27.6} \\
        \addlinespace[2pt]
        MAX         & 82.9 & 43.1 & 88.7 & 51.4 & 91.2 & 51.4 & 64.1 & 27.2 \\
        \quad +RoME & \textbf{83.6} & \textbf{44.4} & \textbf{90.5} & \textbf{53.7} & \textbf{95.2} & \textbf{57.5} & \textbf{65.1} & \textbf{30.5} \\
        \addlinespace[2pt]
        RAMP        & \textbf{83.1} & 43.7 & 89.2 & 52.3 & 90.7 & 55.8 & 62.8 & 29.4 \\
        \quad +RoME & 82.7 & \textbf{46.5} & \textbf{91.3} & \textbf{54.2} & \textbf{93.6} & \textbf{58.7} & \textbf{64.2} & \textbf{32.1} \\
        \bottomrule
    \end{tabular}}
\end{table}

\vspace{0.5ex}\noindent
\textbf{Robust fine-tuning.}
In addition to training the model from scratch, we evaluate our method on robust fine-tuning scenario following E-AT~\cite{e-at}.
For CIFAR-10, we use ViT-B-16 pre-trained with adversarial training on $\ell_\infty$ threats~\cite{pgd} along with WRN-28-10~\cite{wideresnet} and WRN-94-16~\cite{wideresnet} from RobustBench~\cite{robustbench}.
For ImageNet-1K, we use XCiT-S~\cite{xcit} also from RobustBench.
As shown in Table~\ref{tab:finetune}, applying \sys to fine-tuning on $\ell_\infty$-robust models consistently improves union AutoAttack robustness across all models and baseline methods.
For example, applying \sys to RAMP reaches 58.7\% union robustness on WRN-94-16 (+14.0\%p vs. E-AT, +2.9\%p vs. RAMP), with similar trends on WRN-28-10 and XCiT-S.
Beyond robustness, \sys also improves natural accuracy in most settings (\eg, +4.0\%p on WRN-94-16 with MAX).
These consistent gains across ViT, CNN (WRN), and XCiT backbones show that \sys is architecture-agnostic.

\begin{table}[t]
    \centering
    \caption{
    Ablation studies on individual components of \sys on CIFAR-10. 
    $^\ddagger$ indicates using 3 experts for a fair comparison with gating classification, which requires one expert per threat; others use 4 experts. 
    Best results are marked in \textbf{bold}.
    }
    \label{tab:ablation}
    \renewcommand{\arraystretch}{1.05}
    \setlength{\tabcolsep}{8pt} 
    \resizebox{0.9\linewidth}{!}{ 
    \begin{tabular}{l cccccc}
        \toprule
        Methods & Clean & $\ell_1$ & $\ell_2$ & $\ell_\infty$ & Avg. & Union \\
        \midrule
        \rowcolor[gray]{0.95} \multicolumn{7}{l}{\textbf{Low-rank experts (Sec.~\ref{ssec:low_rank_experts})}} \\
        \textit{Expert architectures:} \\
        \hspace{0.5em} FFN-based experts & 76.5 & 46.4 & 58.6 & 31.5 & 45.5 & 30.9 \\
        \hspace{0.5em} Adapter-based experts & 88.0 & 55.2 & \textbf{70.6} & 40.2 & 55.3 & 38.6 \\
        \cmidrule(lr){1-7}
        \hspace{0.5em} \textbf{Ours (Low-rank experts)} & \textbf{88.1} & \textbf{56.4} & 69.7 & \textbf{42.1} & \textbf{56.1} & \textbf{41.1} \\
        \midrule[\heavyrulewidth] 
        
        \rowcolor[gray]{0.95} \multicolumn{7}{l}{\textbf{Dual-scale gating (Sec.~\ref{ssec:bilevel_gating})}} \\
        \textit{No learned gating:} \\
        \hspace{0.5em} No gating (uniform weights) & 87.1 & 53.7 & 69.1 & 38.3 & 53.7 & 38.1 \\
        \cmidrule(lr){1-7}
        \textit{Single-scale gating:} \\
        \hspace{0.5em} Local (token-level) only & 87.5 & 53.1 & 69.6 & 39.9 & 54.2 & 39.4 \\
        \hspace{0.5em} Global (image-level) only & 87.0 & 55.0 & 69.3 & 40.2 & 54.8 & 39.8 \\
        \cmidrule(lr){1-7}
        \textit{Dual-scale gating variants:} \\
        \hspace{0.5em} Inverted layer weights & 87.6 & 55.2 & 69.4 & 39.7 & 54.8 & 39.3 \\
        \hspace{0.5em} Learned layer weights & 87.6 & 53.7 & 69.5 & 40.3 & 54.5 & 39.6 \\
        \hspace{0.5em} CLS token for global & 87.8 & 54.5 & 69.2 & 39.0 & 54.2 & 38.7 \\
        \cmidrule(lr){1-7}
        \hspace{0.5em} \textbf{Ours (avg. + adaptive)} & \textbf{88.1} & \textbf{56.4} & \textbf{69.7} & \textbf{42.1} & \textbf{56.1} & \textbf{41.1} \\
        \midrule[\heavyrulewidth]

        \rowcolor[gray]{0.95} \multicolumn{7}{l}{\textbf{Threat-guided gating diversification loss (Sec.~\ref{ssec:threat_guided})}} \\
        \textit{Conventional MoE regularization:} \\
        \hspace{0.5em} No regularization & 86.7 & 52.9 & 69.3 & 38.7 & 53.6 & 38.1 \\
        \hspace{0.5em} Load balancing~\cite{shazeer2017outrageously} & 86.9 & 55.0 & 69.5 & 38.8 & 54.4 & 38.4 \\
        \hspace{0.5em} Importance~\cite{shazeer2017outrageously} & 87.0 & 53.5 & 69.3 & 39.0 & 53.9 & 38.6 \\
        \cmidrule(lr){1-7}
        \textit{Threat-aware regularization:} \\
        \hspace{0.5em} Gating classification$^\ddagger$ & 87.5 & 42.9 & 69.8 & 34.4 & 49.0 & 31.4 \\
        \cmidrule(lr){1-7}
        \textit{Global gating projection:} \\
        \hspace{0.5em} w/o projection layer & 87.7 & 56.7 & \textbf{69.9} & 41.6 & 56.1 & 40.6 \\
        \cmidrule(lr){1-7}
        \hspace{0.5em} \textbf{Ours (diversification)$^\ddagger$} & 87.9 & \textbf{57.4} & 69.5 & 41.9 & \textbf{56.3} & 41.0 \\
        \hspace{0.5em} \textbf{Ours (diversification)} & \textbf{88.1} & 56.4 & 69.7 & \textbf{42.1} & 56.1 & \textbf{41.1} \\
        \bottomrule
    \end{tabular}}
\end{table}

\subsection{Ablation Studies}
\label{ssec:ablation}

In Table~\ref{tab:ablation}, we provide ablations on individual components of our \sys.

\vspace{0.5ex}\noindent
\textbf{Low-rank experts.}
We verify that FFN experts leads to degraded union robustness, confirming that it is vital to capture threat-common features with a shared backbone.
Our low-rank experts also outperforms adapter~\cite{adapter}-based experts, verifying the effectiveness of our design choices.
In Table~\ref{tab:layers_rome}, we analyze the effect of applying \sys to different layers and find that applying experts to $QKV$ and $O$ projection layers achieves the best efficiency-robustness trade-off.

\vspace{0.5ex}\noindent
\textbf{Dual-scale gating.}
Removing learned gating degrades robustness, showing that threat-specific expert selection is critical.
Single-scale gatings underperform our approach, verifying that different granularities are required for threat discrimination (Fig.~\ref{fig:tsne_dual}).
While inverting or learning the layer-adaptive weights (Eq.~\ref{eq:layer_adaptive}) improves robustness thanks to dual-scale gating, both remain suboptimal, verifying that encoding the Transformer layer hierarchy~\cite{hmora, geva2020transformer} as a fixed prior is more effective than making it learnable.
Replacing averaged global feature with CLS token degrades performance, verifying the effectiveness of our design choices.

\vspace{0.5ex}\noindent
\textbf{Gating diversification.}
Applying no regularization, load balancing~\cite{shazeer2017outrageously}, or importance loss~\cite{shazeer2017outrageously} leads to suboptimal robustness due to threat-agnostic routing.
Gating classification, which trains the gating network as a threat classifier with cross-entropy loss, remains suboptimal as it maps each threat to a single fixed expert unlike our flexible diversification loss.
Removing the projection layer in global diversification reduces union robustness to $40.6\%$, and alternative distance choices in Table~\ref{tab:distance_choice} all underperform our method.
Fig.~\ref{fig:hyperparam} further shows that our method is not overly sensitive to hyperparameters including expert rank, number of experts, loss weight $\lambda$, and $s$, $b$ for layer-adaptive weighting.

\subsection{Analysis on Experts and Threat-Aware Gating}
\label{ssec:analysis_gating}
\begin{figure}[t] 
    \centering
    \begin{minipage}{0.5\textwidth}
        \centering
        \begin{subfigure}{0.48\linewidth}
            \centering
            \includegraphics[width=\linewidth]{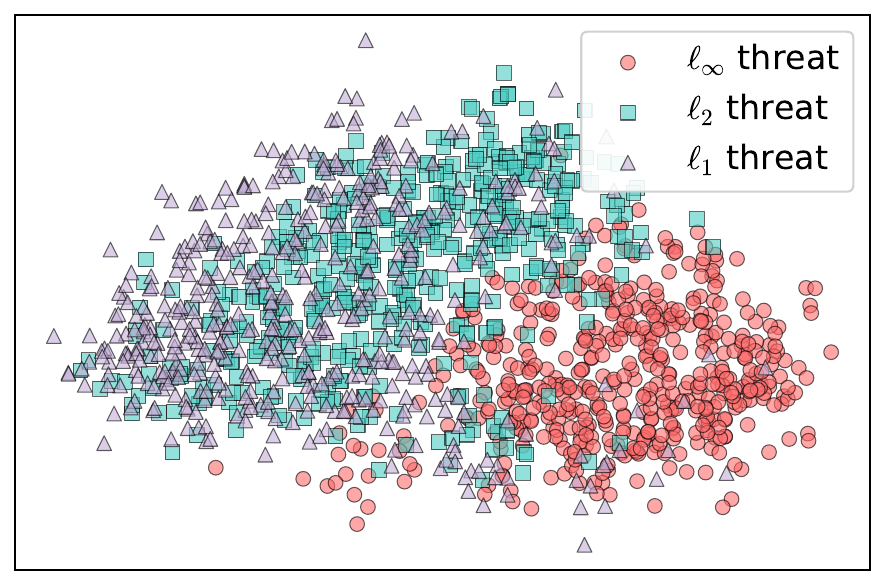}
            \caption{Backbone}
            \label{fig:backbone_update_tsne_c}
        \end{subfigure}
        \hfill
        \begin{subfigure}{0.48\linewidth}
            \centering
            \includegraphics[width=\linewidth]{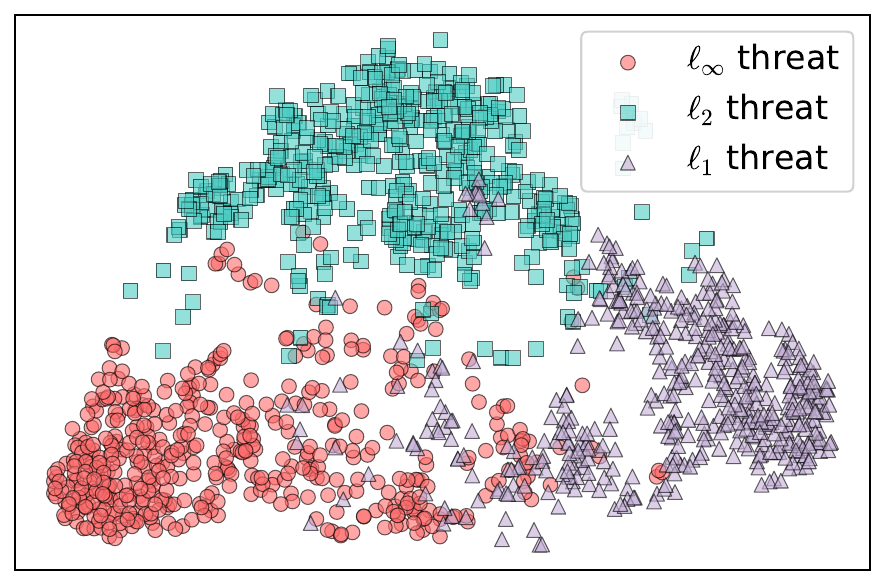}
            \caption{Backbone + experts}
            \label{fig:backbone_update_tsne_d}
        \end{subfigure}
        \caption{
            t-SNE visualization of (a) backbone features and (b) backbone + expert outputs from the final ViT-B layer.
        }
        \label{fig:backbone_update_tsne}
    \end{minipage}
    \hfill 
    \begin{minipage}{0.45\textwidth}
        \vspace{-5pt}
        \centering
        \renewcommand{\arraystretch}{1.2} 
        \setlength{\tabcolsep}{4pt}
        \captionof{table}{
            Analysis on learned experts and gating during inference with RANDOM and PGD training on CIFAR-10.
            Best results are marked in \textbf{bold}.
        }
        \vspace{10pt}
        \label{tab:inference_analysis}
        \resizebox{\textwidth}{!}{
            \begin{tabular}{c|cccccc}
                \specialrule{2pt}{0pt}{2pt}
                Methods & Clean & $\ell_1$ & $\ell_2$ & $\ell_\infty$ & Avg. & Union \\
                \midrule
                Random gating & 88.0 & 55.1 & 69.5 & 39.3 & 54.6 & 38.4 \\
                Inverse gating & 87.8 & 55.4 & 69.0 & 39.6 & 54.7 & 38.9 \\
                w/o experts    & \textbf{88.1} & 54.9 & 69.4 & 39.9 & 54.7 & 39.2 \\
                Ours           & \textbf{88.1} & \textbf{56.4} & \textbf{69.7} & \textbf{42.1} & \textbf{56.1} & \textbf{41.1} \\
                \specialrule{2pt}{2pt}{0pt}
            \end{tabular}
        }
    \end{minipage}
\end{figure}

\vspace{0.5ex}\noindent
\textbf{Learned features on backbone vs. experts.}
To verify whether low-rank experts learn threat-specific features, we analyze t-SNE visualizations of backbone features and expert outputs added to the backbone in Fig.~\ref{fig:backbone_update_tsne}.
Backbone captures similar features regardless of threat type as shown in Fig.~\ref{fig:backbone_update_tsne_c}, showing that it has captured features common among threats.
In contrast, features passed through our low-rank experts capture decoupled representations for each threat as shown in Fig.~\ref{fig:backbone_update_tsne_d}, showing that our experts have learned threat-specific features.

\begin{figure}[t!]
    \centering
    \begin{minipage}{0.5\textwidth}
        \centering
        \begin{subfigure}[b]{\linewidth}
            \centering
            \includegraphics[width=\linewidth]{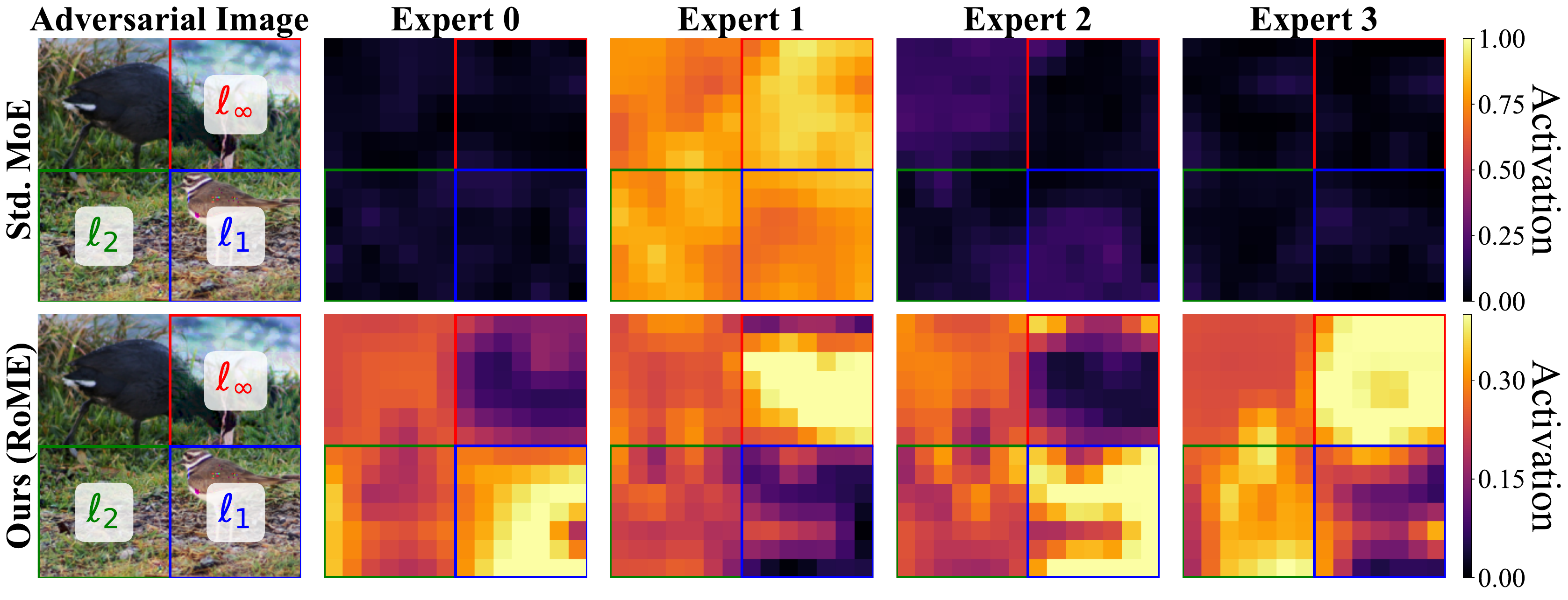}
            \captionsetup{width=\linewidth, format=hang}
            \label{fig:expert_activation_a}
        \end{subfigure}\\
        \vspace{-5pt}
        \begin{subfigure}[b]{\linewidth}
            \centering
            \includegraphics[width=\linewidth]{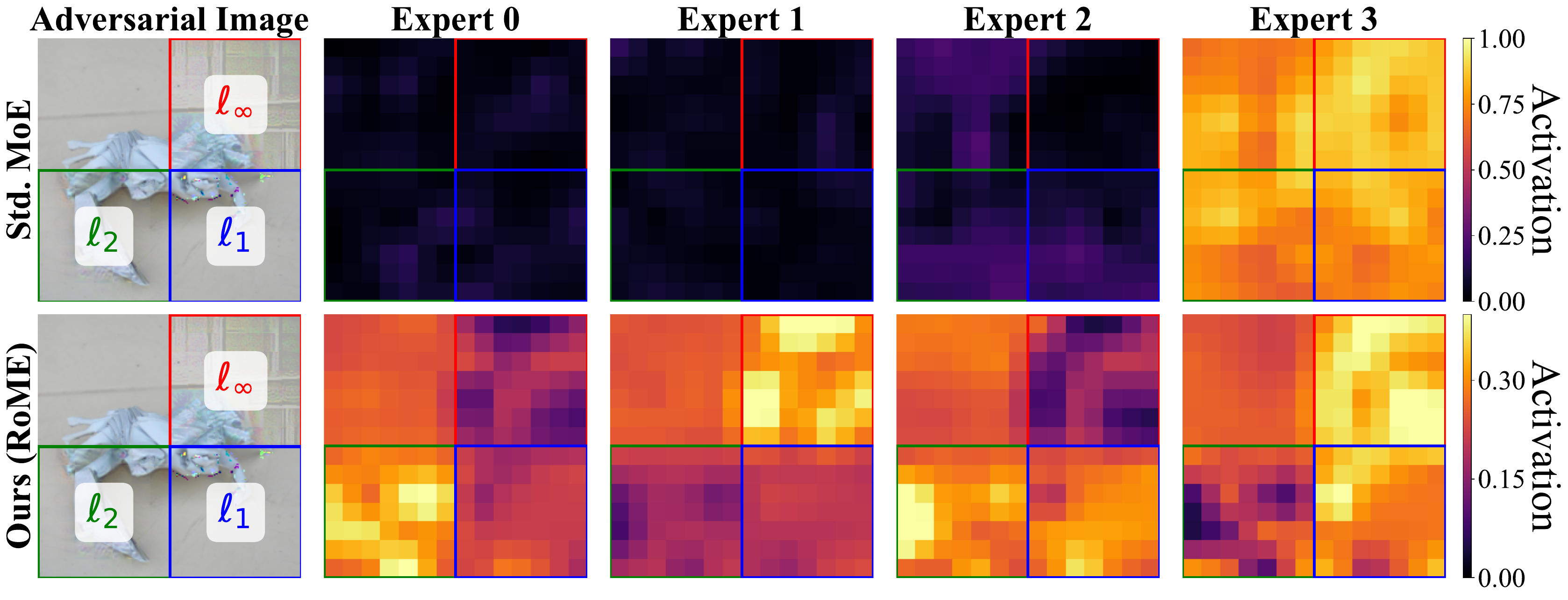}
            \captionsetup{width=\linewidth, format=hang}
            \label{fig:expert_activation_b}
        \end{subfigure}
        \vspace{-15pt}
        \caption{
        Expert activations for various $\ell_p$ threats applied to different regions of image.
        While standard MoE (rows 1, 3) routes threats to same experts, \sys (rows 2, 4) utilizes different experts for each threat.
        }
        \label{fig:expert_activation}
    \end{minipage}
    \hfill
    \begin{minipage}{0.48\textwidth}
        \vspace{-5pt}
        \setlength{\tabcolsep}{7pt}
        \begin{center}
            \centering
            \renewcommand{\arraystretch}{1.15}
            \captionof{table}{
            Computational cost comparison. Best results are marked in \textbf{bold}.
            }
            \vspace{5pt}
            \label{tab:computational_cost}
            \resizebox{\textwidth}{!}{
            \begin{tabular}{c|cccc}
                \specialrule{2pt}{\aboverulesep}{\belowrulesep}
                \multirow{2}{*}{Methods} & \# Params & FLOPs & Train time & Latency \\
                & (M) & (G) & (sec/iter) & (ms/img) \\
                \hline
                RANDOM~\cite{sat} & 85.15 & 5.61 & 3.05 & 0.53 \\
                \hline
                MORE~\cite{more} & 340.60 & 22.43 & 25.16 & 1.70 \\
                MoE (FFN)~\cite{shazeer2017outrageously} & 255.20 & 9.29 & 8.38 & 1.36 \\
                Ours & \textbf{88.85} & \textbf{5.84} & \textbf{3.56} & \textbf{0.60} \\
                \specialrule{2pt}{\aboverulesep}{\belowrulesep}
            \end{tabular}}
        \end{center}


        \begin{center}
            \centering
            \includegraphics[width=0.75\linewidth]{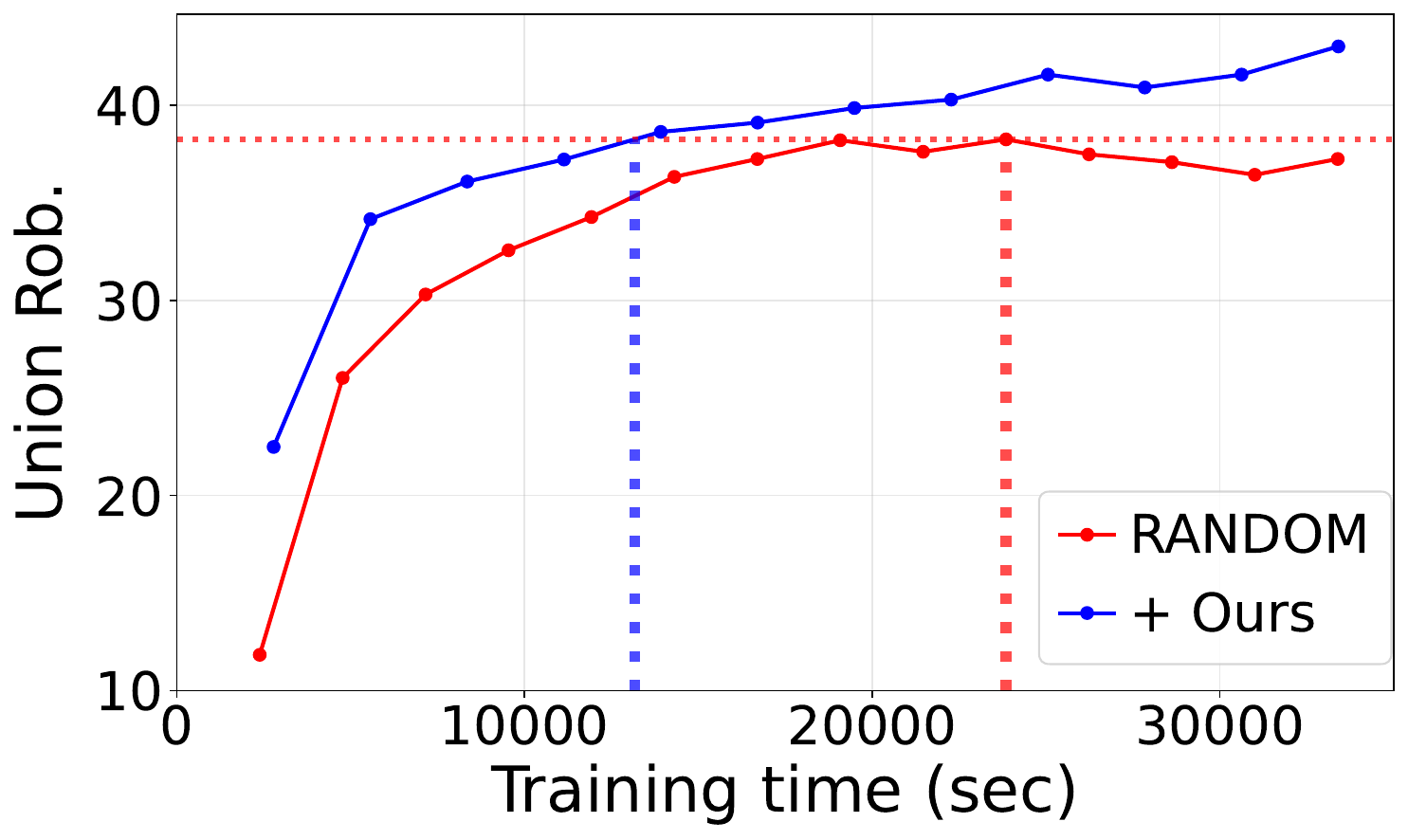}
            \captionof{figure}{
            Training time vs. robustness. Ours consistently outperforms baseline given the same wall-clock training time.
            }
            \label{fig:time_efficiency}
        \end{center}
    \end{minipage}
\end{figure}

\vspace{0.5ex}\noindent
\textbf{Learned routing for $\ell_p$ threats.}
To understand how threats affect expert routing, we apply different threats to various spatial regions and visualize the resulting expert activations for each image patch.
As shown in Fig.~\ref{fig:expert_activation}, standard MoE shows similar expert activations for all image patches (Expert 1 in row 1, left), verifying that experts are not specialized for a specific threat type.
In contrast, our \sys learns threat-specific routing; for example, Expert 1 (row 2) is highly activated for $\ell_\infty$ threats while barely activated for $\ell_1$ threats, demonstrating that each expert specializes in a distinct threat type.

\vspace{0.5ex}\noindent
\textbf{Analysis on experts and gating.}
We analyze learned experts and gating through inference-time variants (Table~\ref{tab:inference_analysis}).
Random weights, inverted gating $1 - \mathbf{g}$, or removing experts degrades robustness, verifying that gating learns threat-specific combinations and that experts learn specialized representations.

\vspace{0.5ex}\noindent
\textbf{Computational analysis.}
In Table~\ref{tab:computational_cost}, we compare computational costs with baseline RANDOM~\cite{sat}.
MORE~\cite{more} and conventional MoE~\cite{shazeer2017outrageously} with FFN experts incur 4$\times$ and 3$\times$ more parameters, with 8.25$\times$ and 2.75$\times$ more train time compared to RANDOM.
In contrast, our method achieves SoTA union robustness (Table~\ref{tab:sota}) with only 1.04$\times$ more parameters and 1.17$\times$ more train time.
Our method also shows superior optimization efficiency, consistently outperforming robustness of RANDOM given the same wall-clock training time (Fig.~\ref{fig:time_efficiency}).
It also reaches RANDOM's peak robustness in 13.2k sec of training versus 23.9k sec for RANDOM.
Baselines with additional expert and gating layers but without our training components in Table~\ref{tab:sota_extra_param} show barely improved robustness, showing that simply increasing model capacity does not address the cross-threat trade-off.

\section{Conclusion}
We introduce \sys, a novel framework that addresses cross-threat robustness trade-off in multi-perturbation adversarial training through mixture of low-rank experts.
We identify the threat-agnostic routing problem where conventional mixture of experts faces difficulty distinguishing between threats and learns similar routing.
To address this, we propose threat-distinguishing dual-scale gating and threat-guided gating diversification that enable distinct expert combinations for each threat. 
Extensive experiments demonstrate that \sys achieves state-of-the-art union robustness across diverse benchmarks, while also improving robustness against a diverse set of unseen threats.
Ablation studies and analysis demonstrate the ability of our method to effectively distinguish between threats and learn diverse threat-specific pathways, leading to improved robustness.

\subsubsection*{\ackname}
Sung-eui Yoon is the corresponding author.
This work was supported by Institute of Information \& communications Technology Planning \& Evaluation (IITP) grant funded by the Korea government (MSIT) (RS-2023-00237965, Recognition, Action and Interaction Algorithms for Open-world Robot Service) and the IITP(Institute of Information \& Communications Technology Planning \& Evaluation)-ITRC(Information Technology Research Center) grant funded by the Korea government(Ministry of Science and ICT)(IITP-2026-RS-2020-II201460).


%
%

\clearpage
\bibliographystyle{splncs04}
\bibliography{main}

\clearpage
\clearpage
\setcounter{page}{1}
\setcounter{table}{0}
\setcounter{section}{0}
\setcounter{figure}{0}

\newcommand{\appendixnumbering}{
    \renewcommand{\thefigure}{A\arabic{figure}}
    \renewcommand{\thetable}{A\arabic{table}}
    \renewcommand{\thesection}{A\arabic{section}}
}

\newpage
\newpage
\appendixnumbering


In this appendix, we first elaborate on additional implementation details (Sec.~\ref{sec:suppl_experiment_setup}).
We report results of applying our \sys to other Transformer-based vision models and ImageNet-1K (Sec.~\ref{sec:suppl_other_models}).
We also provide additional analysis on our \sys (Sec.~\ref{sec:suppl_analysis}), including analysis on gating weights, effects of applying our method on different layers, ablations on our threat-guided gating diversification, and hyperparameter analysis.
Lastly, we provide limitations and future work (Sec.~\ref{sec:suppl_limitations}).

\section{Additional Experimental Setups}
\label{sec:suppl_experiment_setup}
In this section, we elaborate on additional implementation details.

\vspace{0.5ex}\noindent
\textbf{Implementation details.}
The learning rate schedule consists of warmup training for first 4 epochs (1 epoch for ImageNet-100 and ImageNet-1K), linear decay to $0.05$ of initial learning rate for next 12 epochs (3 epochs for ImageNet-100 and ImageNet-1K), then final linear decay to $0.001$ of initial learning rate for the last 4 epochs (1 epoch for ImageNet-100 and ImageNet-1K).
We set batch size to 64, 128, and 128 for CIFAR-10~\cite{cifar10}, ImageNet-100~\cite{imagenet100}, and ImageNet-1K~\cite{imagenet} datasets, respectively.
We set patch size for vision Transformer models to $4$ for CIFAR-10, $16$ for ImageNet-100, and $16$ for ImageNet-1K.

\vspace{0.5ex}\noindent
\textbf{Setup for unseen threats.}
For $\ell_0$ attack, we use SparseFool~\cite{sparsefool} with 10 iterations and $\lambda = 3$.
Following existing benchmarks~\cite{ramp}, we use common corruptions~\cite{common-corruptions} across 5 levels of severity across all corruption types.
For perceptual adversarial threats~\cite{pat}, we set $\epsilon = 12$ for the fog attack, $\epsilon = 0.5$ for the snow attack, $\epsilon = 60$ for the gabor attack, $\epsilon = 0.125$ for the elastic attack, and $\epsilon = 0.125$ for the Jpeg-$\ell_\infty$ attack with $100$ iterations.
For PPGD~\cite{pat} and LPA~\cite{pat}, we use LPIPS bound of $0.25$ with $20$ iterations.
For adversarial patch~\cite{patch}, we use a square shaped patch of size $8 \times 8$ with $50$ iterations and step size of $\frac{2}{255}$.
For spatial StAdv~\cite{stadv} attack, we set flow regularization loss weight $\tau$ to $0.05$.
For semantic ReColorAdv~\cite{recoloradv} attack, we use bound of $0.06$ with $20$ iterations.
For gating misrouting attack (GMA), we combine the original classification loss with a loss that minimizes cross-entropy loss between gating outputs and least-likely expert with the same weight for each loss.

\vspace{0.5ex}\noindent
\textbf{Setup for conventional MoE in Fig.~\ref{fig:gating_diversity}.}
To demonstrate the threat-agnostic issue with conventional mixture of experts when training against multiple adversarial threats, we train ViT-B~\cite{vit} on CIFAR-10~\cite{cifar10}.
We apply experts implemented as FFN layers at each layer of the Transformer block.
We train the entire model using RANDOM strategy~\cite{sat} for 20 epochs with learning rate $\texttt{1e-3}$.

\begin{table}[t]
    \centering
    \renewcommand{\arraystretch}{1.15}
    \caption{
    Comparison with MAT methods using DeiT-B and Swin-B on CIFAR-10 under AutoAttack.
    We measure natural accuracy on clean images, single-threat robustness under $\ell_1$, $\ell_2$, and $\ell_\infty$, their average, and worst-case union robustness.
    Best results are marked in \textbf{bold}.
    }
    \setlength{\tabcolsep}{6pt}
    \resizebox{1.0\linewidth}{!}{
    \begin{tabular}{c|c|cccccc|cccccc}
        \specialrule{2pt}{\aboverulesep}{\belowrulesep}
        \multicolumn{2}{c|}{\multirow{2}{*}{Methods}} & \multicolumn{6}{c|}{\textbf{DeiT-B}} & \multicolumn{6}{c}{\textbf{Swin-B}} \\
        \cmidrule{3-14}
        \multicolumn{2}{c|}{} & Nat & $\ell_1$ & $\ell_2$ & $\ell_\infty$ & Avg & Union & Nat & $\ell_1$ & $\ell_2$ & $\ell_\infty$ & Avg & Union \\
        \midrule
        \multirow{6}{*}{PGD} &
        RANDOM
        & 87.2 & 53.4 & 68.0 & 39.5 & 53.6 & 38.9 & 81.1 & 49.6 & 62.0 & 31.5 & 47.7 & 31.4 \\
        & AVG
        & 86.7 & 52.5 & 67.8 & 39.2 & 53.2 & 38.1 & 80.9 & 49.2 & 61.4 & 31.6 & 47.4 & 31.4 \\
        & MAX
        & 84.8 & 43.1 & 62.5 & 40.3 & 48.6 & 39.5 & 78.5 & 45.1 & 59.7 & 37.1 & 47.3 & 36.2 \\
        & MSD
        & 84.9 & 43.3 & 63.0 & 40.7 & 49.0 & 40.1 & 79.7 & 44.7 & 60.0 & 37.0 & 47.2 & 36.0 \\
        \rowcolor{ourcolor} & RoME+RANDOM (Ours)
        & \textbf{87.6} & \textbf{54.3} & \textbf{70.5} & 40.5 & \textbf{55.1} & 39.7 & \textbf{81.4} & \textbf{51.5} & \textbf{63.4} & 35.0 & \textbf{50.0} & 35.0 \\
        \rowcolor{ourcolor} & RoME+MAX (Ours)
        & 85.4 & 45.3 & 63.9 & \textbf{42.8} & 50.7 & \textbf{41.2} & 79.7 & 47.2 & 60.4 & \textbf{39.2} & 48.9 & \textbf{38.5} \\
        \midrule \midrule
        \multirow{6}{*}{APGD} &
        RANDOM
        & 83.5 & 51.5 & 68.9 & 39.6 & 53.3 & 38.7 & 77.6 & 47.4 & 62.5 & 31.9 & 47.3 & 31.3 \\
        & MAX
        & 79.3 & 45.8 & 64.1 & 43.9 & 51.3 & 41.9 & 73.1 & 42.8 & 58.5 & 36.2 & 45.8 & 35.6 \\
        & E-AT
        & 80.5 & 52.1 & 68.3 & 40.2 & 53.5 & 39.0 & 74.5 & 47.7 & 62.1 & 32.1 & 47.3 & 31.5 \\
        & RAMP
        & 79.7 & 46.9 & 63.8 & 44.0 & 51.6 & 42.3 & 74.2 & 43.3 & 58.8 & 37.1 & 46.4 & 36.1 \\
        \rowcolor{ourcolor} & RoME+RANDOM (Ours)
        & \textbf{84.5} & \textbf{53.9} & \textbf{70.3} & 40.6 & \textbf{54.9} & 39.9 & \textbf{78.9} & \textbf{50.0} & \textbf{63.1} & 34.7 & \textbf{49.3} & 34.6 \\
        \rowcolor{ourcolor} & RoME+MAX (Ours)
        & 80.1 & 49.0 & 65.4 & \textbf{44.3} & 52.9 & \textbf{43.3} & 75.1 & 44.6 & 60.3 & \textbf{39.8} & 48.2 & \textbf{39.3} \\
        \specialrule{2pt}{\aboverulesep}{\belowrulesep}
    \end{tabular}}
    \label{tab:sota_deit}
\end{table}

\vspace{0.5ex}\noindent
\textbf{Setup for feature analysis in Fig.~\ref{fig:tsne_dual}.}
To analyze feature separability across different scales, we train ViT-B~\cite{vit} without any experts using RANDOM strategy~\cite{sat} on CIFAR-10~\cite{cifar10}.
We train the model using RANDOM strategy~\cite{sat} for 20 epochs with learning rate $\texttt{1e-3}$.

\begin{table}[t]
    \centering
    \renewcommand{\arraystretch}{1.15}
    \setlength{\tabcolsep}{6pt}
    \caption{
    Comparison with MAT methods using ViT-B on ImageNet-1K under AutoAttack.
    We measure natural accuracy on clean images, single-threat robustness under $\ell_1$, $\ell_2$, and $\ell_\infty$, their average, and worst-case union robustness.
    Best results are marked in \textbf{bold}.
    }
    \resizebox{0.7\linewidth}{!}{
    \begin{tabular}{c|c|cccccc}
        \specialrule{2pt}{\aboverulesep}{\belowrulesep}
        \multicolumn{2}{c|}{Methods} & Nat & $\ell_1$ & $\ell_2$ & $\ell_\infty$ & Avg & Union \\
        \midrule
        &
        RANDOM
        & 64.9 & 18.6 & 41.3 & 20.1 & 26.7 & 15.3 \\
        & MAX
        & 60.5 & 19.4 & 39.4 & 22.3 & 27.0 & 18.6 \\
        \rowcolor{ourcolor} & RoME+RANDOM (Ours)
        & \textbf{66.2} & 20.2 & \textbf{42.4} & 22.0 & 28.2 & 17.0 \\
        \rowcolor{ourcolor} \multirow{-4}{*}{PGD} & RoME+MAX (Ours)
        & 61.6 & \textbf{21.4} & 39.6 & \textbf{24.7} & \textbf{28.6} & \textbf{20.3} \\
        \midrule \midrule
        &
        RANDOM
        & 63.3 & 19.5 & 41.8 & 20.5 & 27.3 & 15.9 \\
        & MAX
        & 60.3 & 20.0 & 39.1 & 23.1 & 27.4 & 19.0 \\
        \rowcolor{ourcolor} & RoME+RANDOM (Ours)
        & \textbf{63.8} & 20.9 & \textbf{42.8} & 22.8 & 28.8 & 17.2 \\
        \rowcolor{ourcolor} \multirow{-4}{*}{APGD} & RoME+MAX (Ours)
        & 61.2 & \textbf{21.6} & 40.3 & \textbf{25.3} & \textbf{29.1} & \textbf{20.6} \\
        \specialrule{2pt}{\aboverulesep}{\belowrulesep}
    \end{tabular}}
    \label{tab:sota_imagenet}
\end{table}
\begin{table*}[t]
    \centering
    \renewcommand{\arraystretch}{1.15}
    \caption{
    Comparison with baselines trained on the same number of parameters as our method by simply adding expert and gating network layers without our proposed components.
    We use ViT-B as backbone model.
    Best results are marked in \textbf{bold}.
    }
    \setlength{\tabcolsep}{6pt}
    \resizebox{0.9\textwidth}{!}{
    \begin{tabular}{c|c|cccccc|c}
        \specialrule{2pt}{\aboverulesep}{\belowrulesep}
        & Methods & Nat & $\ell_1$ & $\ell_2$ & $\ell_\infty$ & Avg & Union & \# Params \\
        \midrule
        \multirow{6}{*}{PGD}
        & RANDOM 
        & 86.7 & 52.3 & 68.8 & 40.3 & 53.8 & 38.6 & 85.15  \\
        & + experts \& gating layers
        & 86.9 & 52.4 & 68.8 & 40.4 & 53.9 & 38.6 & 88.85  \\
        & MAX
        & 85.1 & 46.6 & 66.8 & 40.1 & 51.2 & 38.5 & 85.15 \\
        & + experts \& gating layers
        & 84.9 & 46.7 & 66.7 & 40.1 & 51.2 & 38.5 & 88.85  \\
        \rowcolor{ourcolor} & RoME+RANDOM (Ours)
        & \textbf{88.1} & \textbf{56.4} & \textbf{69.7} & 42.1 & \textbf{56.1} & 41.1 & 88.85 \\
        \rowcolor{ourcolor} & RoME+MAX (Ours)
        & 85.4 & 48.2 & 68.2 & \textbf{43.1} & 53.2 & \textbf{42.3} & 88.85 \\
        \midrule \midrule
        \multirow{6}{*}{APGD}
        & RANDOM 
        & 86.5 & 50.1 & 69.3 & 37.5 & 52.3 & 37.2 & 85.15 \\
        & + experts \& gating layers
        & 85.5 & 50.8 & 69.3 & 37.6 & 52.6 & 37.3 & 88.85  \\
        & MAX
        & 81.7 & 45.5 & 65.3 & 42.6 & 51.1 & 41.4 & 85.15 \\
        & + experts \& gating layers
        & 82.3 & 45.1 & 65.5 & 42.2 & 50.9 & 41.2 & 88.85  \\
        \rowcolor{ourcolor} & RoME+RANDOM (Ours)
        & \textbf{87.4} & \textbf{54.7} & \textbf{71.6} & 40.0 & \textbf{55.4} & 39.4 & 88.85  \\
        \rowcolor{ourcolor} & RoME+MAX (Ours)
        & 82.5 & 48.9 & 67.3 & \textbf{44.2} & 53.5 & \textbf{43.7} & 88.85  \\
        \specialrule{2pt}{\aboverulesep}{\belowrulesep}
    \end{tabular}}
    \label{tab:sota_extra_param}
\end{table*}

\begin{figure}[!t]
    \centering
    \begin{subfigure}[b]{0.7\linewidth}
        \centering
        \includegraphics[width=\linewidth]{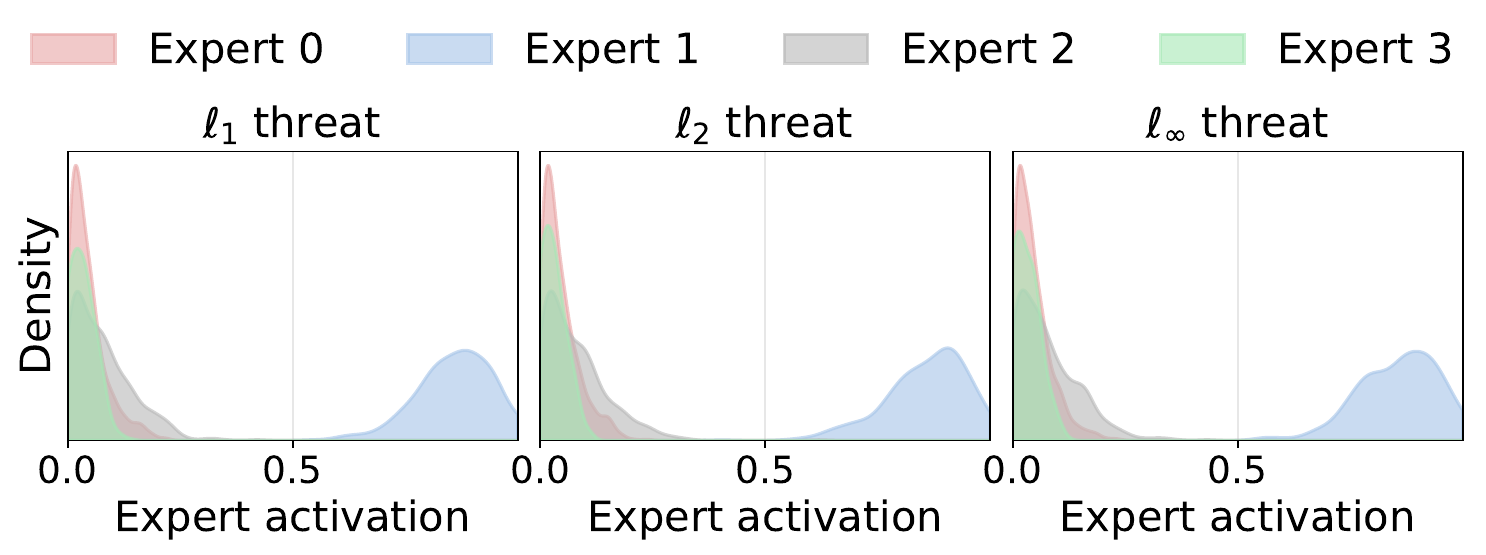}
        \captionsetup{width=\linewidth, format=hang}
        \caption{Na\"ive approach}
        \label{fig:gating_diversity_naive_suppl}
    \end{subfigure}
    \hfill
    \begin{subfigure}[b]{0.7\linewidth}
        \centering
        \includegraphics[width=\linewidth]{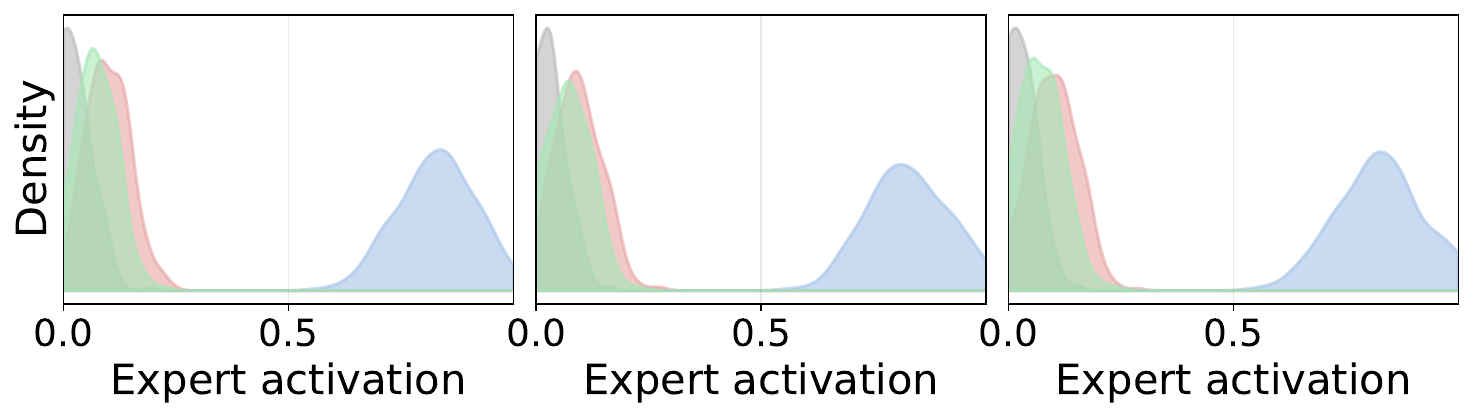}
        \captionsetup{width=\linewidth, format=hang}
        \caption{Load balancing loss~\cite{shazeer2017outrageously} ($\lambda_\text{lb} = 0.00005$)}
        \label{fig:gating_diversity_loadbal_suppl_low}
    \end{subfigure}
    \hfill
    \begin{subfigure}[b]{0.7\linewidth}
        \centering
        \includegraphics[width=\linewidth]{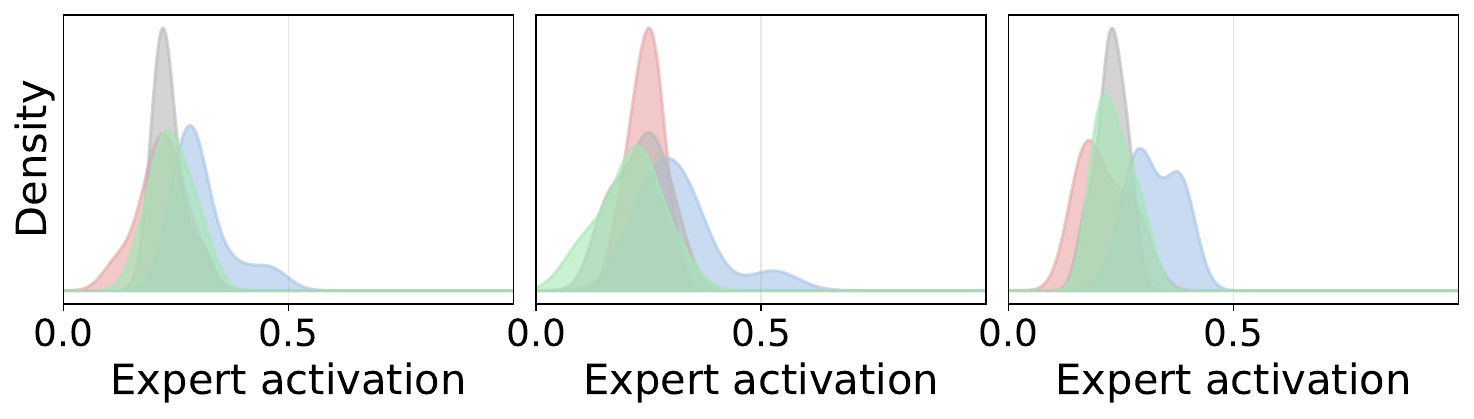}
        \captionsetup{width=\linewidth, format=hang}
        \caption{Load balancing loss~\cite{shazeer2017outrageously} ($\lambda_\text{lb} = 0.005$)}
        \label{fig:gating_diversity_loadbal_suppl_med}
    \end{subfigure}
    \hfill
    \begin{subfigure}[b]{0.7\linewidth}
        \centering
        \includegraphics[width=\linewidth]{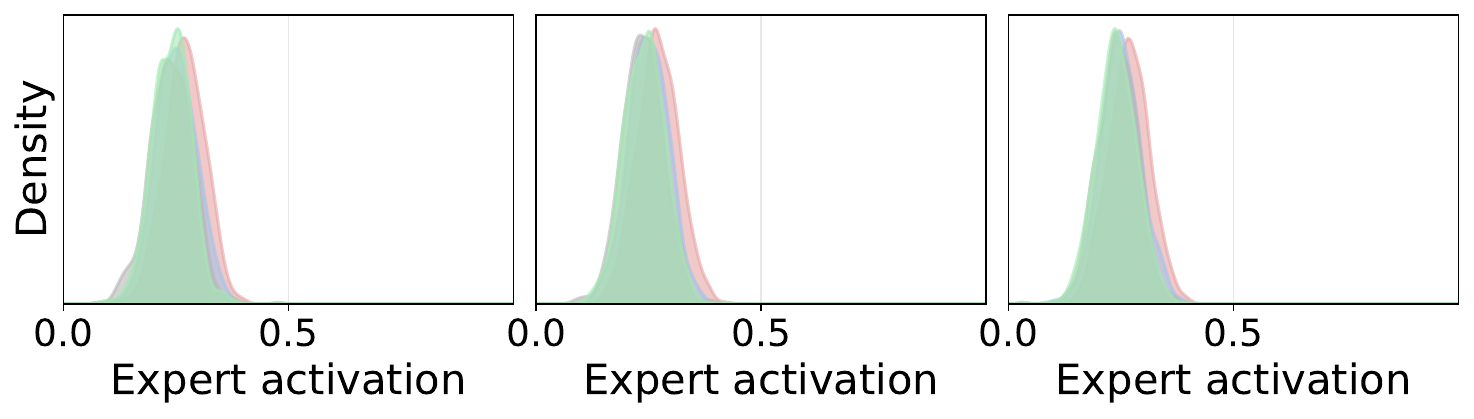}
        \captionsetup{width=\linewidth, format=hang}
        \caption{Load balancing loss~\cite{shazeer2017outrageously} ($\lambda_\text{lb} = 0.5$)}
        \label{fig:gating_diversity_loadbal_suppl_high}
    \end{subfigure}
    \hfill
    \begin{subfigure}[b]{0.7\linewidth}
        \centering
        \includegraphics[width=\linewidth]{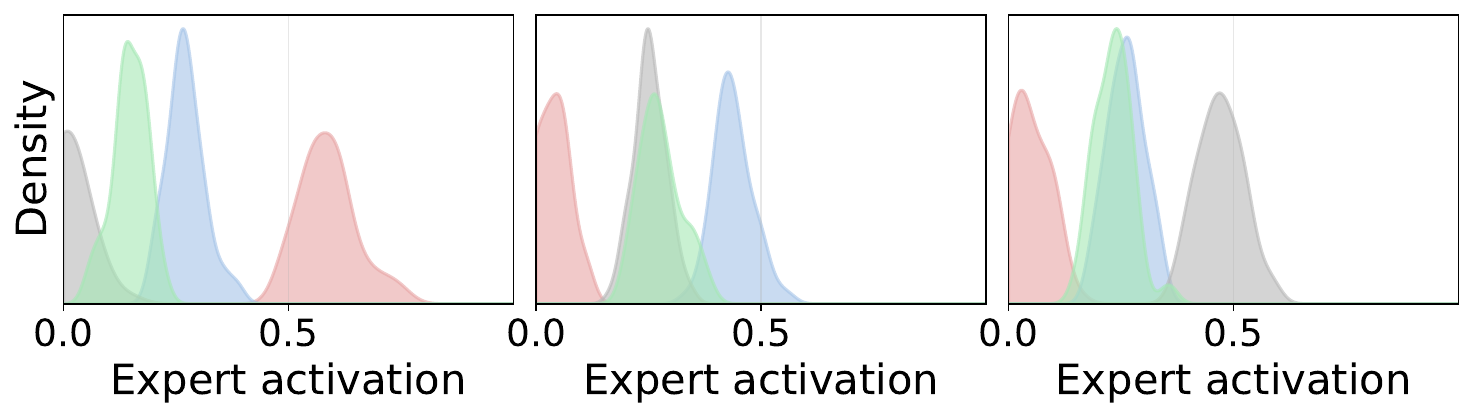}
        \captionsetup{width=\linewidth, format=hang}
        \caption{Ours (\sys)}
        \label{fig:gating_diversity_ours_suppl}
    \end{subfigure}
    \caption{
    Analysis on gating weight distributions across different threat types on CIFAR-10 using Gaussian KDE.
    (a) Na\"ive MoE training suffers from both routing collapse (Expert 1 dominates) and threat-agnostic routing (similar distributions across threats). 
    (b-d) Load balancing~\cite{shazeer2017outrageously} resolves routing collapse but still exhibits threat-agnostic routing regardless of its loss coefficient $\lambda_\text{lb}$.
    (e) Our \sys framework learns diverse, threat-specific expert combinations.
    }
    \label{fig:gating_diversity_lam}
\end{figure}

\section{Application on Other Models and Dataset}
\label{sec:suppl_other_models}
\vspace{0.5ex}\noindent
\textbf{Deit-B and Swin-B.}
In Table~\ref{tab:sota_deit}, we report comparison of our method with previous multi-perturbation adversarial training methods under DeiT-B~\cite{deit} and Swin-B~\cite{swin} on CIFAR-10.
Our approach using MAX outperforms the union robustness of existing methods by at least 1.1\%p under PGD, and our approach using RANDOM outperforms the average robustness and natural accuracy of existing methods by at least 1.5\%p and 0.3\%p, respectively, under PGD.

\vspace{0.5ex}\noindent
\textbf{ImageNet-1K dataset.}
In Table~\ref{tab:sota_imagenet}, we evaluate our method on the large-scale ImageNet-1K~\cite{imagenet} dataset.
Applying \sys improves union robustness over the baseline RANDOM and MAX methods by 1.7\%p and 1.7\%p under PGD, and by 1.3\%p and 1.6\%p under APGD, respectively.
These consistent gains even on a large-scale dataset verify the scalability and generalizability of our \sys.

\vspace{0.5ex}\noindent
\textbf{Baselines with more parameters.}
In Table~\ref{tab:sota_extra_param}, we compare with baselines trained on the same model architectures as our method where extra layers for experts and gating networks are applied, but without our core training components.
We observe that simply increasing the model capacity from 85.15M to 88.85M parameters by adding experts and gating layers results in negligible performance gains.
For example, adding these layers to the RANDOM strategy under APGD training only marginally improves union robustness from 37.2\% to 37.3\%, which remains lower than the 39.4\% achieved by our RoME+RANDOM.
In some instances, such as the MAX strategy under APGD, the additional layers without our proposed components even lead to a slight degradation in union robustness from 41.4\% to 41.2\%. 
These findings demonstrate that simply increasing the network capacity with extra expert and gating network layers does not address the robustness trade-offs in multi-perturbation adversarial training.
Instead, the effectiveness of \sys stems specifically from our dual-scale gating and threat-guided gating diversification, which provide the necessary discriminative signals and supervision to enable diverse threat-specific model pathways.

\section{Additional Analysis}
\label{sec:suppl_analysis}

\begin{figure} [t]
    \centering
    \includegraphics[width=\linewidth]{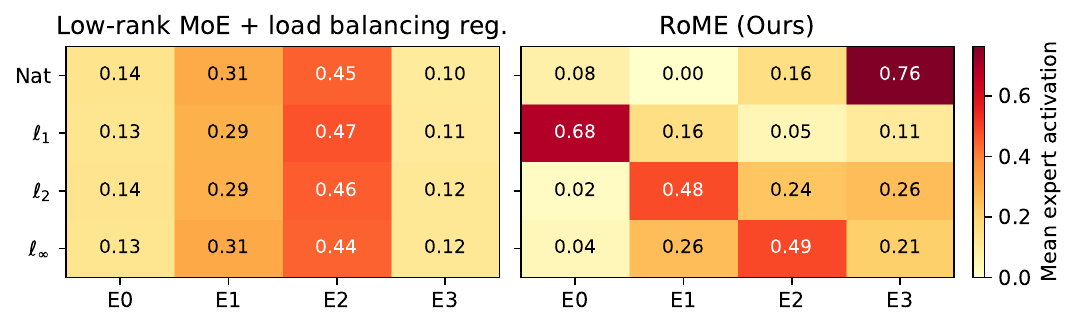}
    \caption{
    Gating activation visualization of low-rank mixture of experts with load balancing loss (left) and our \sys with dual-scale gating and threat-guided gating diversification (right).
    While baseline MoE fails to distinguish between threats and assigns the same expert combinations to different threats, our method effectively decouples model pathways for natural images and different threats.
    }
    \label{fig:gating_activations}
\end{figure}

\subsection{Gating Weights Similarity}
In addition to Fig.~\ref{fig:gating_diversity} in the main paper, we visualize gating weight distributions for different loss coefficients $\lambda_\text{lb}$ of the load balancing loss~\cite{shazeer2017outrageously} in Fig.~\ref{fig:gating_diversity_lam}.
As $\lambda_\text{lb}$ decreases, the gating distribution (Fig.~\ref{fig:gating_diversity_loadbal_suppl_low}) degenerates to the na\"ive approach, thus heavily relying on the same single expert for all threats.
As $\lambda_\text{lb}$ increases, the gating distribution (Fig.~\ref{fig:gating_diversity_loadbal_suppl_high}) becomes more uniform throughout the expert, thus making gating weights similar for all threats.
Either way, load balancing loss fails to address the threat-agnostic issue and highlights the need for our \sys.

\subsection{Gating Activations}
\label{ssec:gating_activations}
In Fig.~\ref{fig:gating_activations}, we visualize the gating activations for each experts.
Simply applying low-rank mixture of experts and load balancing loss (left) shows that the gating network fails to distinguish between threats and collapses to using the same expert combinations for all threats.
In contrast, our \sys with dual-scale gating and threat-guided gating diversification (right) allocates different expert combinations for each threat, effectively specializing each expert for different threats.
This allows mitigation of cross-threat trade-off, thus improving union robustness.

This allocation of experts for different threats also provides insight on improvements in natural accuracy compared to baseline methods (Sec.~\ref{ssec:main_results}).
As shown in Fig.~\ref{fig:gating_activations} (right), our \sys assigns a unique combination of experts for natural images, thus separating the model pathways for natural and adversarial images.
This prevents a single model pathway from capturing both natural accuracy and adversarial robustness, thus alleviating the well-known accuracy-robustness trade-off in adversarial training~\cite{trades, mart} and thus improving natural accuracy.

\begin{table*}[t]
    \centering
    \renewcommand{\arraystretch}{1.15}
    \caption{
    Analysis on \sys application across different layers (QKV, output projection, and FFN) in ViT-B-16~\cite{vit}.
    \# Params denotes additional parameters incurred with mixture of low-rank experts.
    Best results are marked in \textbf{bold}.
    }
    \setlength{\tabcolsep}{6pt}
    \resizebox{\linewidth}{!}{
    \begin{tabular}{ccc|cccccc|c}
        \specialrule{2pt}{\aboverulesep}{\belowrulesep}
        \multicolumn{3}{c|}{Layers} 
        & \multirow{2}{*}{Clean}
        & \multirow{2}{*}{$\ell_1$}
        & \multirow{2}{*}{$\ell_2$}
        & \multirow{2}{*}{$\ell_\infty$}
        & \multirow{2}{*}{Avg.}
        & \multirow{2}{*}{Union} 
        & \multirow{2}{*}{\# Params} \\
        \cline{1-3}
        \textbf{Attn-QKV} & \textbf{Attn-Out} & \textbf{FFN} 
        &  &  &  &  &  &  & \\
        \hline
        \checkmark & \texttimes & \texttimes & 87.4 & 53.7 & 70.4 & 41.4 & 55.2 & 41.0 & 2.44M \\
        \texttimes & \checkmark & \texttimes & 87.6 & 53.3 & 71.4 & 40.8 & 55.2 & 39.9 & \textbf{1.26M} \\
        \texttimes & \texttimes & \checkmark & 86.7 & 54.8 & 71.1 & 41.2 & 55.7 & 40.5 & 6.28M \\
        \checkmark & \checkmark & \texttimes & \textbf{88.1} & \textbf{56.4} & 69.7 & 42.1 & 56.1 & 41.1 & 3.70M \\
        \checkmark & \texttimes & \checkmark & 86.9 & 55.2 & \textbf{71.8} & 39.8 & 55.6 & 39.3 & 8.82M \\
        \texttimes & \checkmark & \checkmark & 87.5 & 55.7 & 71.3 & 41.9 & \textbf{56.3} & 41.0 & 7.54M \\
        \checkmark & \checkmark & \checkmark & 86.4 & 55.9 & 70.1 & \textbf{42.5} & 56.2 & \textbf{41.9} & 9.98M \\
        \specialrule{2pt}{\aboverulesep}{\belowrulesep}
    \end{tabular}}
    \label{tab:layers_rome}
\end{table*}
\begin{table}[t]
    \centering
    \renewcommand{\arraystretch}{1.15}
    \caption{
    Analysis on varying distance choices for threat-guided gating diversification with RANDOM and PGD training on CIFAR-10.
    Best results are marked in \textbf{bold}.
    }
    \setlength{\tabcolsep}{5pt}
    \resizebox{0.75\columnwidth}{!}{
    \begin{tabular}{c|cccccc}
        \specialrule{2pt}{\aboverulesep}{\belowrulesep}
        Methods & Clean & $\ell_1$ & $\ell_2$ & $\ell_\infty$ & Avg. & Union \\
        \hline
        No regularization       & 86.7 & 52.9 & 69.3 & 38.7 & 53.6 & 38.1 \\
        Cosine similarity       & 87.0 & 54.8 & 69.6 & 40.2 & 54.9 & 39.4 \\
        KL divergence           & 87.5 & 55.7 & \textbf{70.5} & 39.8 & 55.3 & 39.4 \\
        Wasserstein distance    & 87.3 & 55.4 & 70.3 & 40.8 & 55.5 & 40.0 \\
        Ours ($L$2)             & \textbf{88.1} & \textbf{56.4} & 69.7 & \textbf{42.1} & \textbf{56.1} & \textbf{41.1} \\
        \specialrule{2pt}{\aboverulesep}{\belowrulesep}
    \end{tabular}}
    \label{tab:distance_choice}
\end{table}

\subsection{Analysis on Layers of \sys Application}
We apply our low-rank experts on different layer types of the backbone model in ViT-B-16.
As shown in Table~\ref{tab:layers_rome}, applying experts to all layers -- query/key/value projection (Attn-QKV), output projection (Attn-Out), and feed-forward networks (FFN) -- achieves the highest robustness.
However, this comes at a greater computational cost.
We find that applying \sys to query, key, value, and output projection layers achieves comparable performance with fewer parameters, providing a better efficiency-robustness trade-off.

\subsection{Different Distance Functions for Threat-Guided Gating Diversification}
We replace the $L2$ distance function in Eqs.~\ref{eq:local_div} and~\ref{eq:global_div} with different distance functions in Table~\ref{tab:distance_choice}.
We first replace it with cosine similarity (row 2), which maximizes the cosine similarity between the two gating weights to maximize angular separation in the expert usage space.
We also replace it with KL divergence (row 3), which maximizes the symmetric KL divergence (\ie, $\text{KL}(P||Q) + \text{KL}(Q||P)$) between gating distributions to encourage divergent probability mass allocations across experts.
We also try Wasserstein distance (row 4), which maximizes the 1-Wasserstein distance (earth mover's distance) computed as the $L1$ distance between cumulative distribution functions of sorted gating scores.
All three variants outperform no regularization (row 1), verifying the necessity of learning different gating weights for each threat.
However, they underperform our $L2$ distance.

\begin{figure*}[t]
    \centering
    \begin{subfigure}[b]{0.32\linewidth}
        \centering
        \includegraphics[width=\linewidth]{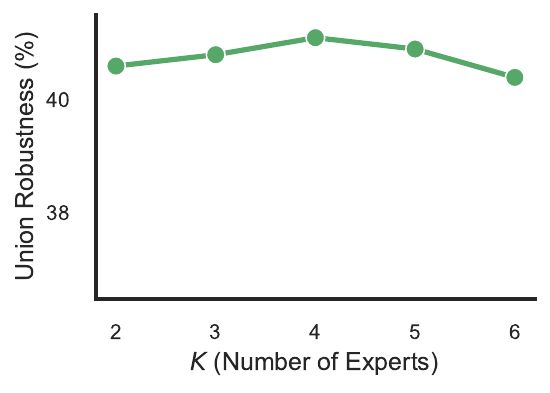}
        \caption{Number of experts $K$}
        \label{fig:hyperparam_num_experts}
    \end{subfigure}
    \hfill
    \begin{subfigure}[b]{0.32\linewidth}
        \centering
        \includegraphics[width=\linewidth]{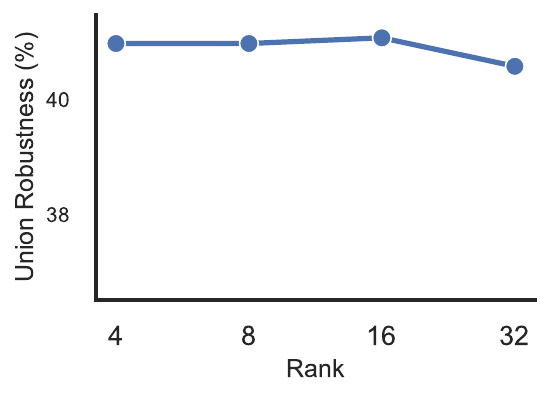}
        \caption{Rank for experts}
        \label{fig:hyperparam_rank}
    \end{subfigure}
    \hfill
    \begin{subfigure}[b]{0.32\linewidth}
        \centering
        \includegraphics[width=\linewidth]{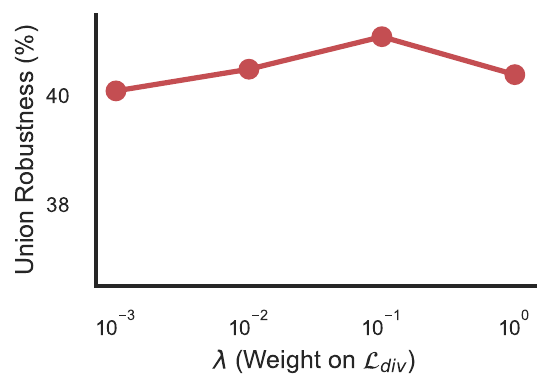}
        \caption{$\lambda$ for $\mathcal{L}_\text{div}$}
        \label{fig:hyperparam_lam_gate}
    \end{subfigure}\\
    \begin{subfigure}[b]{0.32\linewidth}
        \centering
        \includegraphics[width=\linewidth]{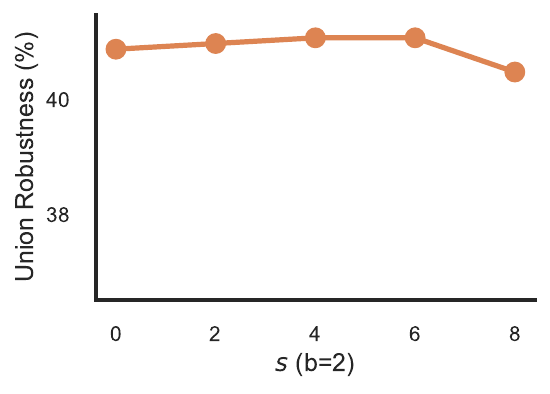}
        \caption{$s$ for layer-adaptive coefficient}
        \label{fig:hyperparam_s}
    \end{subfigure}
    \begin{subfigure}[b]{0.32\linewidth}
        \centering
        \includegraphics[width=\linewidth]{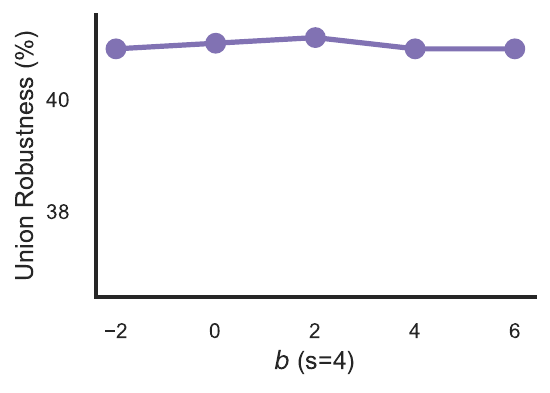}
        \caption{$b$ for layer-adaptive coefficient}
        \label{fig:hyperparam_b}
    \end{subfigure}
    \caption{
    Effects of varying hyperparameters (a) number of experts, (b) rank for experts, (c) $\lambda$ for threat-guided gating diversification $\mathcal{L}_\text{div}$, and (d-e) $s$ and $b$ for layer-adaptive coefficient on union robustness with RANDOM and PGD training on CIFAR-10.
    }
    \label{fig:hyperparam}
\end{figure*}

\subsection{Hyperparameter Analysis}
\label{ssec:hyperparam}
In Fig.~\ref{fig:hyperparam}, we provide sensitivity analysis on various hyperparameters.

\vspace{0.5ex}\noindent
\textbf{Number and capacity of experts.}
As shown in Figs.~\ref{fig:hyperparam_num_experts} and ~\ref{fig:hyperparam_rank}, we analyze on various number of experts $K$ and the rank of individual low-rank experts.
Union robustness improves as $K$ increases from 2 to 4, showing that it is vital to capture diverse characteristics of threats with multiple experts, while the performance degrades as $K$ increases beyond 4.
We can observe similar behavior of improved robustness when increasing the rank to $16$, but degraded performance as we increase it to $32$.
This is a common phenomenon in mixture of low-rank experts~\cite{lorahub, mole, loramoe} where increasing the computational capacity of experts after a certain amount degrades performance.

\vspace{0.5ex}\noindent
\textbf{Gating diversification loss weight $\lambda$.}
As shown in Fig.~\ref{fig:hyperparam_lam_gate}, we analyze on varying weight $\lambda$ for threat-guided gating diversification $\mathcal{L}_\text{div}$ (Eq.~\ref{eq:total_div}).
In general, higher weight leads to higher union robustness, verifying the importance of our gating diversification loss.
Too high $\lambda$ slightly degrades robustness.
This is because excessively large $\lambda$ creates a strong conflict between the classification objective and the diversification objective, causing the gating network to prioritize threat-guided gating diversification over task performance.

\vspace{0.5ex}\noindent
\textbf{Layer-adaptive coefficient $s$ and $b$.}
As shown in Figs.~\ref{fig:hyperparam_s} and~\ref{fig:hyperparam_b}, we analyze on varying values of $s$ and $b$ that control the layer-adaptive coefficient $\beta(l)$ (Eq.~\ref{eq:layer_adaptive}).
This schedule balances global and local gating to exploit the architectural hierarchy of Transformers.
As shown in Fig.~\ref{fig:hyperparam_s}, union robustness peaks at $s=4$, suggesting that a moderately steep transition between global and local signals is optimal.
Excessively high values (e.g., $s=8$) cause a drop in robustness by forcing a premature transition.
Similarly as shown in Fig.~\ref{fig:hyperparam_b}, the shift parameter $b$ achieves its highest performance at $b=2$, with deviations in either direction leading to suboptimal results.

\begin{figure}[!t]
    \centering
    \textbf{Layer 0}\\[0.5em]
    \begin{subfigure}[b]{0.24\linewidth}
        \centering
        \includegraphics[width=\linewidth]{fig/tsne_local_a.pdf}
        \caption{First patch}
    \end{subfigure}
    \hfill
    \begin{subfigure}[b]{0.24\linewidth}
        \centering
        \includegraphics[width=\linewidth]{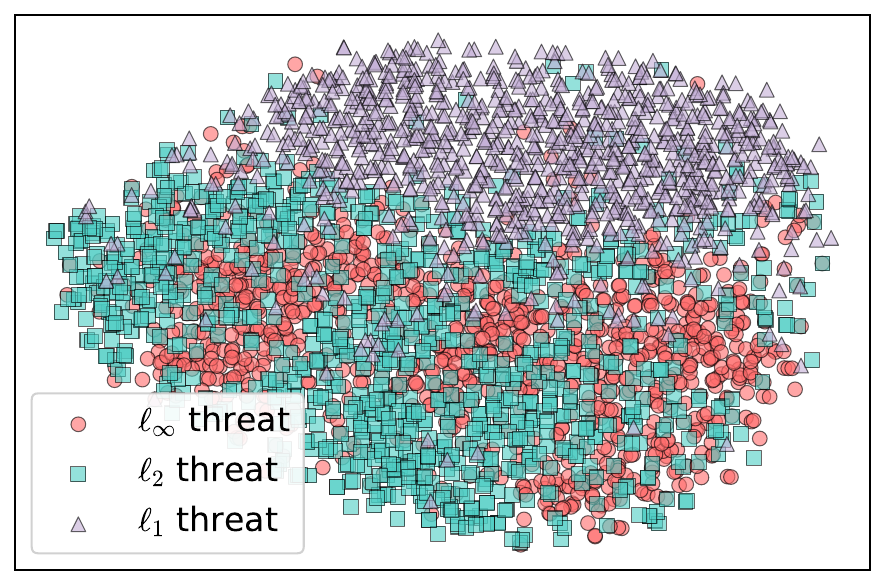}
        \caption{Middle patch}
    \end{subfigure}
    \hfill
    \begin{subfigure}[b]{0.24\linewidth}
        \centering
        \includegraphics[width=\linewidth]{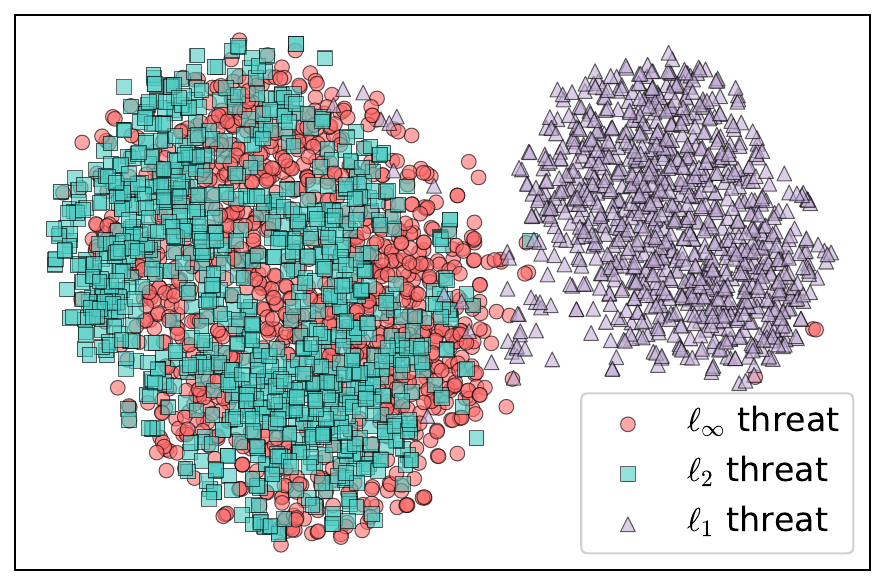}
        \caption{Last patch}
    \end{subfigure}
    \hfill
    \begin{subfigure}[b]{0.24\linewidth}
        \centering
        \includegraphics[width=\linewidth]{fig/tsne_global.pdf}
        \caption{Global}
    \end{subfigure}
    
    \vspace{1.5em}
    
    \textbf{Layer 11}\\[0.5em]
    \begin{subfigure}[b]{0.24\linewidth}
        \centering
        \includegraphics[width=\linewidth]{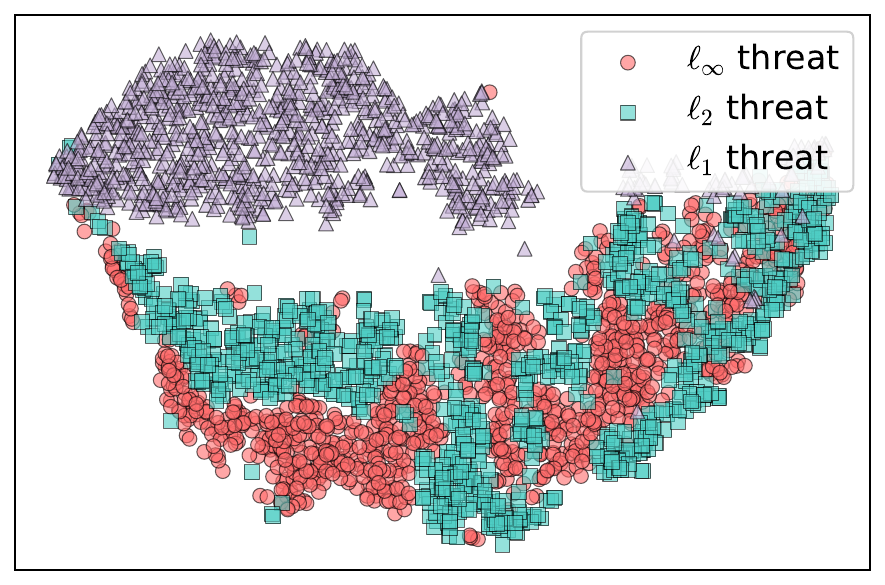}
        \caption{First patch}
    \end{subfigure}
    \hfill
    \begin{subfigure}[b]{0.24\linewidth}
        \centering
        \includegraphics[width=\linewidth]{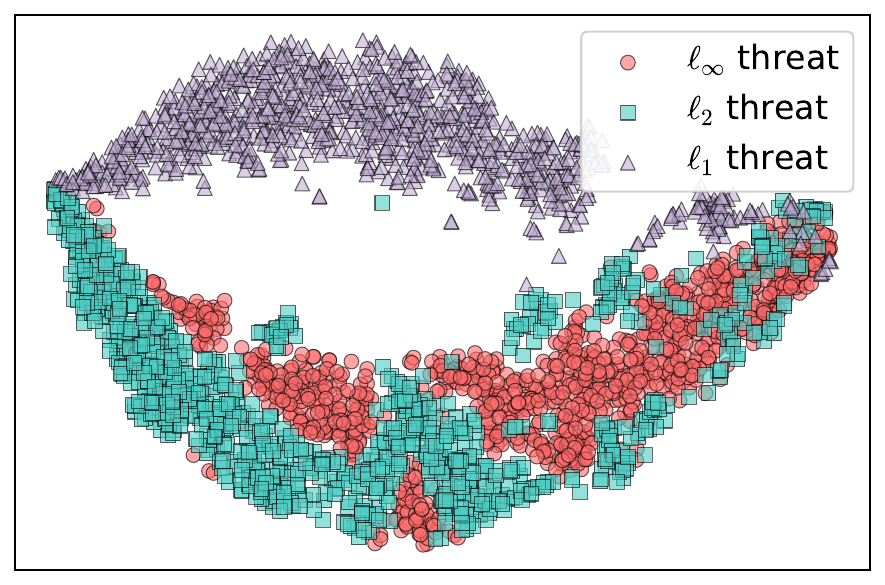}
        \caption{Middle patch}
    \end{subfigure}
    \hfill
    \begin{subfigure}[b]{0.24\linewidth}
        \centering
        \includegraphics[width=\linewidth]{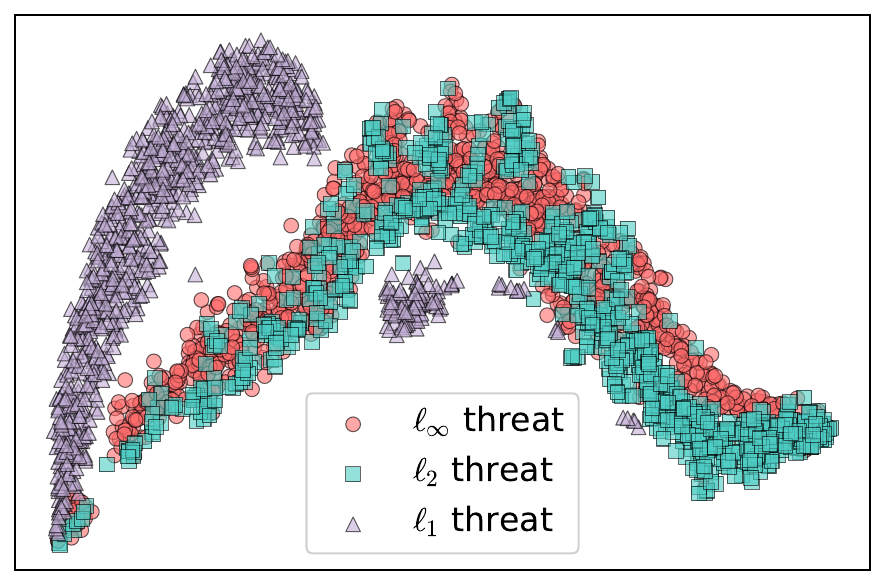}
        \caption{Last patch}
    \end{subfigure}
    \hfill
    \begin{subfigure}[b]{0.24\linewidth}
        \centering
        \includegraphics[width=\linewidth]{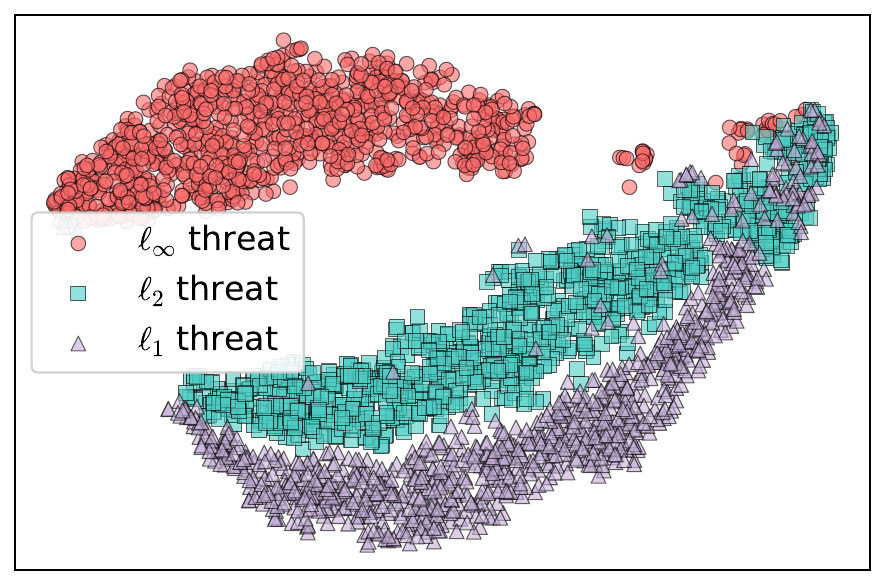}
        \caption{Global}
    \end{subfigure}
    
    \caption{
    t-SNE visualization of threat separability of local patch-level features at token first, middle, and last token positions and the global-level features at layer 0 (top) and layer 11 (bottom).
    }
    \label{fig:tsne_all_patch}
\end{figure}

\subsection{Additional Analysis on Dual-Scale Feature Separability}
\label{ssec:dual_scale_analysis}
In Fig.~\ref{fig:tsne_all_patch}, we show t-SNE visualization for multiple patch-level features in addition to Fig.~\ref{fig:tsne_dual}.
Specifically, we visualize features from three representative patches: the first, middle, and last patches, and global feature.
We also visualize feature separability at layer 0 (top) and 11 (bottom) of ViT-B.
Local patch-level features consistently show difficulty separating $\ell_\infty$ threats, while global image-level features better distinguish $\ell_\infty$ threats, demonstrating the need for our dual-scale gating strategy.

\section{Limitations and Future Work}
\label{sec:suppl_limitations}
Although our method has shown promising results, we train it on a fixed set of threats.
It thus requires fully updating the experts and gating networks when training on new threat types after deployment.
Future work could explore continual adaptation such as progressive expert addition~\cite{pnn, pmoe} or dynamic expansion~\cite{ddas} to incrementally expand threat coverage for long-term deployment scenarios~\cite{crt} without retraining the entire model.

\end{document}